\documentclass[letterpaper]{article} 
\usepackage{aaai2026}  
\usepackage{times}  
\usepackage{helvet}  
\usepackage{courier}  
\usepackage[hyphens]{url}  
\usepackage{graphicx} 
\usepackage{natbib}  
\usepackage{caption} 
\usepackage{parskip} 
\usepackage{algorithm}
\usepackage{algorithmic}

\usepackage{amsmath}
\usepackage{multirow}
\usepackage{booktabs}
\usepackage{amssymb}

\usepackage{newfloat}
\usepackage{listings}
\DeclareCaptionStyle{ruled}{labelfont=normalfont,labelsep=colon,strut=off} 
\floatstyle{ruled}
\newfloat{listing}{tb}{lst}{}
\floatname{listing}{Listing}
\title{LungNoduleAgent: A Collaborative Multi-Agent System for Precision Diagnosis of Lung Nodules}
\author{
    Cheng Yang\textsuperscript{\rm 1}, Hui Jin\textsuperscript{\rm 1}, Xinlei Yu\textsuperscript{\rm 2}, Zhipeng Wang\textsuperscript{\rm 1}, Yaoqun Liu\textsuperscript{\rm 3}, Fenglei Fan\textsuperscript{\rm 4},\\ Dajiang Lei\textsuperscript{\rm 5}, Gangyong Jia\textsuperscript{\rm 1}, Changmiao Wang\textsuperscript{\rm 6}, Ruiquan Ge\textsuperscript{\rm 1,}\thanks{Corresponding author: Ruiquan Ge, gespring@hdu.edu.cn.}
}
\affiliations{
    \textsuperscript{\rm 1}  Hangzhou Dianzi University, Hangzhou, China\\
    \textsuperscript{\rm 2} National University of Singapore, Singapore\\
    \textsuperscript{\rm 3} The Chinese University of Hong Kong, Shenzhen, Shenzhen, China\\
    \textsuperscript{\rm 4} City University of Hong Kong, Hong Kong, China\\
    \textsuperscript{\rm 5} Chongqing University of Posts and Telecommunications, Chongqing, China\\
    \textsuperscript{\rm 6} Shenzhen Research Institute of Big Data, Shenzhen, China
}

\usepackage{bibentry}

\begin{document}

\maketitle

\begin{abstract}
Diagnosing lung cancer typically involves physicians identifying lung nodules in Computed tomography (CT) scans and generating diagnostic reports based on their morphological features and medical expertise. Although advancements have been made in using multimodal large language models for analyzing lung CT scans, challenges remain in accurately describing nodule morphology and incorporating medical expertise. These limitations affect the reliability and effectiveness of these models in clinical settings. Collaborative multi-agent systems offer a promising strategy for achieving a balance between generality and precision in medical applications, yet their potential in pathology has not been thoroughly explored. To bridge these gaps, we introduce LungNoduleAgent, an innovative collaborative multi-agent system specifically designed for analyzing lung CT scans. LungNoduleAgent streamlines the diagnostic process into sequential components, improving precision in describing nodules and grading malignancy through three primary modules. The first module, the Nodule Spotter, coordinates clinical detection models to accurately identify nodules. The second module, the Radiologist, integrates localized image description techniques to produce comprehensive CT reports. Finally, the Doctor Agent System performs malignancy reasoning by using images and CT reports, supported by a pathology knowledge base and a multi-agent system framework. Extensive testing on two private datasets and the public LIDC-IDRI dataset indicates that LungNoduleAgent surpasses mainstream vision-language models, agent systems, and advanced expert models such as GPT-4o, Claude 3.7 Sonnet, LLaMA-3.2 Vision, Qwen2.5-VL, Med-R1, MedGemma, MedAgent-Pro, MedAgents, MDAgent and LLaVA-Med. These results highlight the importance of region-level semantic alignment and multi-agent collaboration in diagnosing nodules. LungNoduleAgent stands out as a promising foundational tool for supporting clinical analyses of lung nodules. 
\end{abstract}

\begin{links}
    \link{Code}{https://github.com/ImYangC7/LungNoduleAgent}
\end{links}

\section{Introduction}
Lung cancer remains a leading cause of cancer-related deaths globally, with early detection and accurate diagnosis being vital for improving patient outcomes \cite{tammemagi2019predicting}. Computed tomography (CT) scans are crucial for identifying lung nodules, which serve as early indicators of malignancy \cite{swanson2023patterns}. Traditionally, radiologists examine these scans by evaluating nodule morphology and applying their medical expertise to produce diagnostic reports \cite{osarogiagbon2023evaluation}. However, this process necessitates radiologists to manually examine each subsequent CT image \cite{hammer2019decision,lee2024read}, which is time-consuming and susceptible to interobserver variability \cite{driessen2025completeness}.

\begin{figure*}[t]
\centering
\includegraphics[width=1.9\columnwidth]{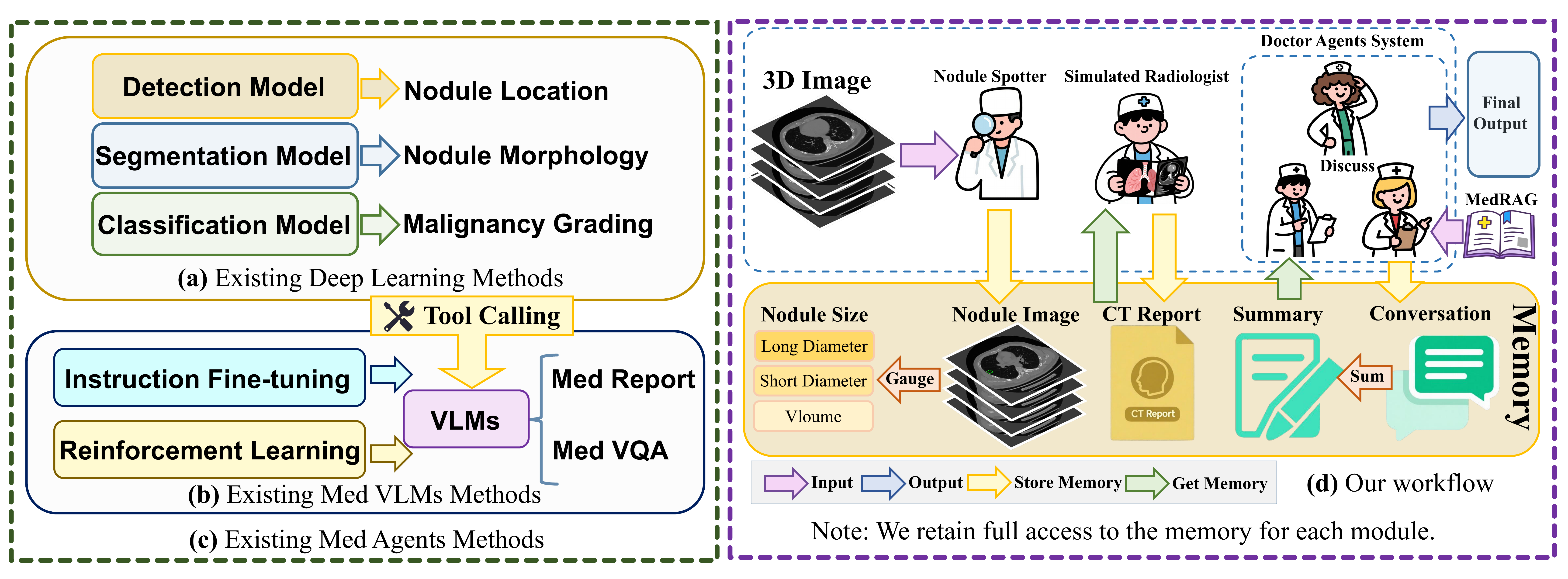}
\caption{Workflow Comparison of Traditional Methods vs. the Proposed LungNoduleAgent Framework.}
\label{fig:1}
\end{figure*}

In the realm of lung cancer analysis using CT images, deep learning advancements have greatly improved the screening and diagnosis of nodules \cite{swanson2023patterns}. These technological strides primarily focus on three key tasks: cancer classification \cite{nakach2024comprehensive, raza2023lung, ji2023lung, yu2025small}, grading \cite{fan2025deep, shen2025csf, martinez2021deep, wang2018automated}, and nodule detection \cite{cao2020two, urrehman2024effective}. Despite these achievements, deep learning models still face significant challenges \cite{ladbury2023integration}. One major issue is the interpretability of model outputs. While these models often achieve high performance metrics, their reasoning processes are not easily understood, which impedes their acceptance in clinical settings. Unlike traditional clinical reasoning, which follows transparent diagnostic procedures that physicians can explain to patients, deep learning models typically offer conclusions without revealing the specific features or patterns that underpin their predictions. Consequently, this lack of transparency poses a barrier to their integration into clinical practice. Another challenge is the dependency of deep learning models on diverse datasets. When confronted with new, unseen data, these models may fail to generalize effectively. This limitation highlights the necessity for models that are adaptable to various scenarios. Furthermore, many current methods are designed for specific tasks, reducing their flexibility. This situation underscores the urgent need for comprehensive models capable of addressing multiple pathological tasks while also providing enhanced interpretability.

Recently, newly released general Vision-Language Models (VLMs) \cite{zhang2024mm} such as LLaVA \cite{liu2023llava}, GPT-4o \cite{achiam2023gpt}, Qwen2.5VL \cite{Qwen2.5-VL}, InternVL \cite{zhu2025internvl3} and Claude 3.7 Sonnet \cite{anthropic2024claude} have shown impressive interpretation and generalization abilities. However, these models often fall short in specialized medical contexts due to insufficient domain-specific training and a lack of domain prior knowledge, which prevents them from meeting the demands of professional medical applications.
Medical VLMs~\cite{thirunavukarasu2023large}, including Med-R1~\cite{lai2025med}, LLaVA-Med~\cite{li2023llavamed}, PMC-VQA~\cite{zhang2023pmc}, and MedGemma~\cite{sellergren2025medgemma}, which are based on supervised fine-tuning and reinforcement learning, have improved their multimodal reasoning and generalization capabilities in the medical field. Nevertheless, their limited fine-grained visual perception hinders quantitative analysis. These models often rely on the internal knowledge of VLMs for judgments, while modern medical practice emphasizes evidence-based diagnosis, which requires structured reasoning and clinical evidence.
Simultaneously, agentic systems~\cite{liu2023agentbench,yang2025multi,yu2025visual,yu2025snow} that use collaborative multi-agent systems~\cite{liang2023encouraging,du2023improving,chan2023chateval} for clinical reasoning or integrate external DL tools~\cite{wang2025medagent,li2024mmedagent,fallahpour2025medrax} to extend the capabilities of VLMs offer a promising solution. However, their application in lung cancer diagnosis is not yet mature, lacking fine-grained analysis of lung nodules and sufficient pathology-specific knowledge. This leads to an accuracy of only 40-50\% on lung cancer-specific tasks, while achieving 75-80\%~\cite{tang2023medagents,kim2024mdagents,wang2024beyond,liu2024medchain} on general medical tasks. This underscores the need for a more comprehensive and collaborative diagnostic framework in the field of nodule analysis \cite{qiu2025embodied}.

To address these challenges, we introduce LungNoduleAgent, the first collaborative multi-agent system for lung nodule analysis, designed to enhance the accuracy and reliability of lung CT scan analysis. As shown in \textbf{Figure \ref{fig:1}}, our contributions are as follows:

\begin{itemize}
    \item The system mimics the clinical workflow by decomposing the diagnostic process into sequential stages, each handled by a specialized component:  ``Nodule Spotter" focuses on precise nodule detection,  ``Simulated Radiologist" generates detailed CT reports, and  ``Doctor Agent System (DAS)" assesses malignancy by leveraging expert knowledge and multi-agent discussions. 
    \item We employ a Focal Prompting Mechanism to enhance the model's fine-grained visual perception, enabling the description of nodule morphological features and the dynamic evaluation of nodule characteristics across slices. 
    \item By storing pathological information in a shared Memory for the agent system, we utilize collaborative multi-agent discussions and introduce medical prior knowledge to achieve evidence-based quantitative analysis and fact-based clinical reasoning.

\end{itemize}
Comprehensive evaluations on two private and one public dataset show our system is significantly more effective than existing VLMs and medical agentic systems in nodule report generation and malignancy assessment.

\begin{figure*}[t]
\centering
\includegraphics[width=1.6\columnwidth]{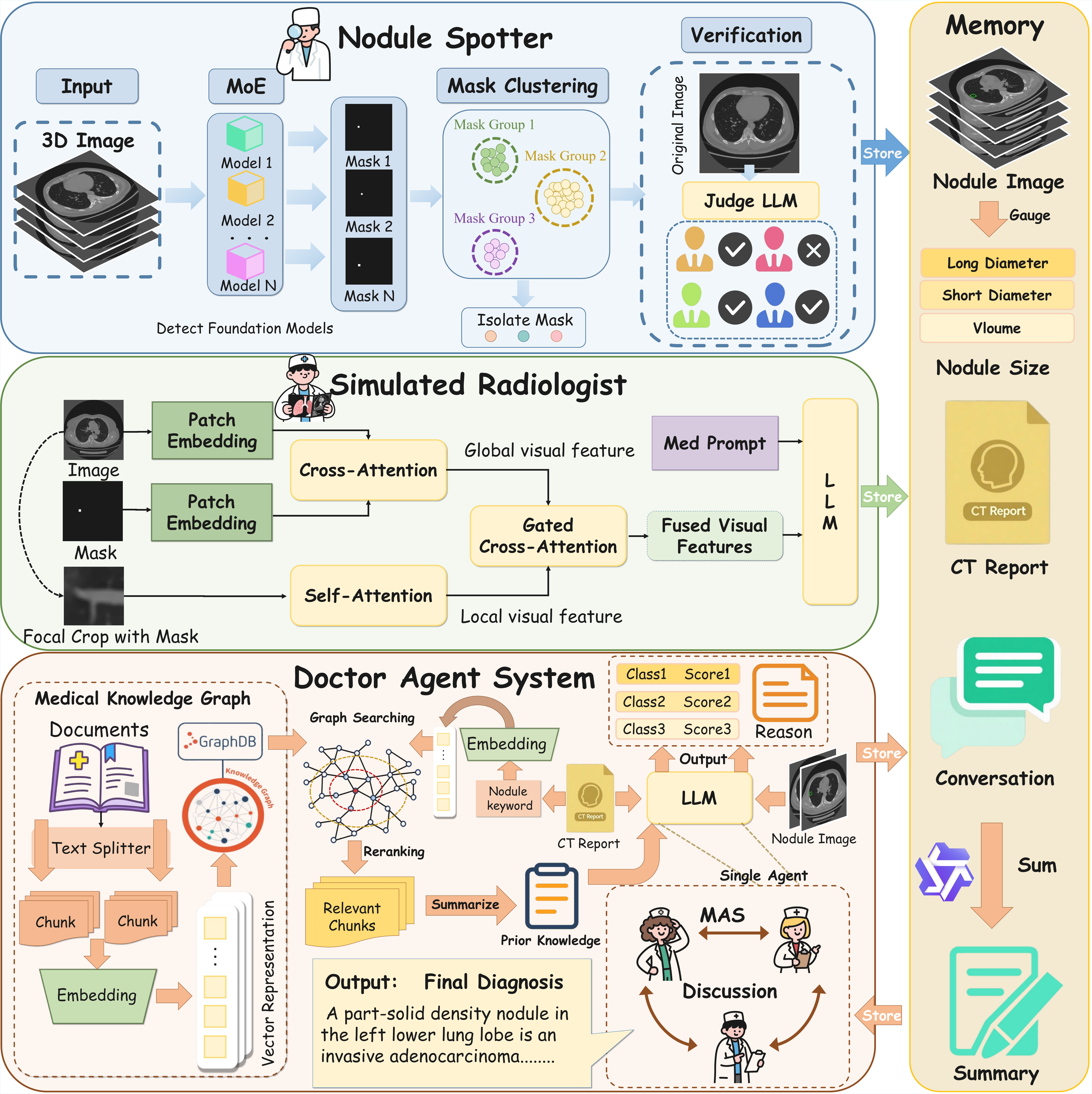}
\caption{Overview of LungNoduleAgent for multi-modal lung nodule analysis with a nodule spotter for lung nodule detection, simulated radiologist for localized CT report generation, and Doctor Agent System for malignancy grading.}
\label{fig:2}
\end{figure*}

\section{Methodology}
\subsection{Overview Architecture}
The workflow of LungNoduleAgent, illustrated in \textbf{Figure \ref{fig:2}}, processes lung CT volume \(V\) , through three  modules: Nodule Spotter, Simulated Radiologist, Doctor Agent System. 

The Nodule Spotter begins by dividing the CT volume into slices \(I\), which are then processed through a Mixture of Experts architecture to produce initial masks \(m\). These masks go through Mask Clustering, which refines them into candidate masks \(M_g\) and excludes outliers. A Judging Panel uses VLMs to validate \(M_g\), determining the final mask \(M\).

In the Simulated Radiologist module, the image slice \(I\) and the final mask \(M\) are encoded using a Focal Prompting Mechanism that highlights lung nodule regions while retaining surrounding context. A MedPrompt further refines the generation of CT reports, ensuring focus on annotated areas and the integration of visual and textual information to produce localized CT Reports. Both the Nodule Image and the CT Report are stored in the Memory module for future access and interaction.

The Doctor Agent System employs each Medical Agent as a VLM enhanced with a medical knowledge graph. This integration of domain-specific medical knowledge enriches the VLMs with relevant pathological information, aiding in constructing reasoning chains and providing diagnostic explanations. Initially, each agent, with unique domain expertise, analyzes the Nodule Image and CT Report to generate its diagnosis and supporting rationale, stored in the Memory's Conversation module to generate. A designated VLM then compiles these initial outputs into a summary, which is used for a second round of discussion among the agents. This iterative process continues until consensus is reached, culminating in the final diagnosis $\mathcal{FD}$.

\subsection{Nodule Spotter} 
The Nodule Spotter module is essential in the diagnostic process, as it identifies and pinpoints lung nodule regions. This capability allows radiologists to create descriptions that are tailored to these localized findings.

\subsubsection{Mixture of Experts.} This submodule processes 3D images via slice-based division for analysis. It adopts a Mixture of Experts (MoE) framework, incorporating multiple specialized foundational detection models, each proficient in identifying specific nodule types or features. These models operate in parallel, analyzing individual lung CT slices independently, with each expert generating a \(m\) to denote predicted nodule locations. Such parallel processing leverages the strengths of diverse models, enhancing the robustness and comprehensiveness of initial nodule detection through the integration of their outputs.

\subsubsection{Mask Clustering.}
This module clusters initial masks based on their spatial overlap to obtain a more stable and accurate final mask representation. Masks with substantial overlap are grouped into the same cluster, while those with insufficient similarity to any group are treated as outliers and removed.

To quantify mask similarity, we define the distance between two masks \(m_i\) and \(m_j\) as
\begin{equation}
d(m_i, m_j) = 1 - \mathrm{IoU}(m_i, m_j),
\end{equation}
where the Intersection over Union is
\begin{equation}
\mathrm{IoU}(m_i, m_j) = \frac{|m_i \cap m_j|}{|m_i \cup m_j|},
\end{equation}
and \(|\cdot|\) denotes the number of pixels. Distances therefore lie in \([0,1]\), with smaller values indicating stronger spatial overlap. This metric forms the basis for mask grouping.

We adopt DBSCAN for clustering, parameterized by a neighborhood radius \(\epsilon \in [0,1]\)---equivalent to an IoU threshold \(\tau = 1 - \epsilon\)---and a minimum sample count \(\mathrm{MinPts}\). The \(\epsilon\)-neighborhood of a mask is defined as
\begin{equation}
N_\epsilon(m_i) = \{ m_j \mid d(m_i, m_j) \le \epsilon \}.
\end{equation}
DBSCAN marks all masks as unvisited initially. For each mask \(m_i\), if \(|N_\epsilon(m_i)| \ge \mathrm{MinPts}\), a new cluster is created and expanded by recursively adding all masks reachable through connected \(\epsilon\)-neighborhoods. Masks that do not meet this density requirement are labeled noise. The algorithm outputs clusters \(C = \{C_1, \ldots, C_K\}\) and noise set \(N\).

For each cluster \(C_k = \{ m_1, \ldots, m_N \}\), we compute an averaged mask \(\bar{m}_k\) to aggregate the spatial consensus among its members. The averaged mask is then binarized with a threshold of 0.5, producing the final refined mask used for subsequent nodule analysis.

\subsubsection{Judging Panel.}
The Judging Panel validation submodule is the crucial final step in distinguishing true nodule candidates from false positives, utilizing a consensus approach among VLMs. For each nodule candidate, represented by the mask group \(M_g\) and the original image slice \(I\), \(N_{VLM}\) independent VLMs simultaneously assess the visual alignment between the mask and anatomical features. After undergoing evaluation by the Judging Panel, the final mask \(M\) is determined.

Each VLM, denoted as \(\mathcal{V}_j\), performs its assessment through two outputs: (1) a binary decision, \(\text{Sign}(\mathcal{V}_j)\), where \(+1\) indicates approval as a valid nodule, and \(-1\) indicates rejection, as defined by:
\begin{equation}
\text{Sign}(\mathcal{V}_j) = 
\begin{cases} 
+1, &  \text{approval}, \\
-1, & \text{rejection},
\end{cases}
\end{equation}
and (2) a confidence score \(C_j\) that quantifies the certainty of the alignment.

The system consolidates these evaluations through weighted scoring:
\begin{equation}
\text{Score}(M_g) = \sum_{j=1}^{N_{VLM}} (\text{Sign}(\mathcal{V}_j) \times C_j),
\end{equation}
where a \(\text{Score}(M_g) > 0\) confirms the validity of the candidate through majority consensus.

This design, combining discrete decisions with continuous confidence metrics, simulates a rigorous peer-review process. It mitigates individual model biases while establishing interpretable thresholds suitable for clinical applications. Nodule candidates deemed invalid, with \(\text{TotalScore} \leq 0\), are automatically filtered out as false positives, ensuring that only anatomically accurate detections are considered for further analysis.

\subsection{Simulated Radiologist}
\subsubsection{Focal Prompting Mechanism.} Simulated Radiologist utilizes a focal prompting technique inspired by the Describe Anything Model \cite{zhang2021dual} to encode areas with lung nodules while preserving the surrounding context for thorough analysis. This technique involves focal cropping of both the image and its corresponding mask, ensuring that nearby areas are maintained for local context.





Both the full image and the focal crop are processed by a local visual backbone, embedding the image and binary mask in a spatially aligned manner. To deepen the understanding of the focal crop, global context from the full image is integrated through gated cross-attention. Through sequence concatenation, the dynamic changes of nodules across slices are captured to enhance the overall perception of the nodules \cite{mao2025ct}. This approach enhances the detailed comprehension and analysis of the nodule regions within the broader anatomical context. 

\subsubsection{Region-specific Prompting.}

We employ MedPrompt to guide the effective generation of CT reports. Unlike general prompts, MedPrompt ensures that the model focuses exclusively on annotated areas, utilizing anatomically accurate terminology and maintaining a professional clinical report format while avoiding speculative or irrelevant content. It incorporates clearly defined output formats and customized objectives to reduce errors and inaccurate descriptions. 
 The fused feature, combined with the prompt, is then processed by VLM:
 \begin{equation}
 \mathcal{O}_{vlm} = \text{VLM}(\text{MedPrompt},\Theta_{\text{volume}}).
 \end{equation}
This process ensures the production of accurate and clinically relevant CT reports $\mathcal{O}_{\text{vlm}}$.

\begin{figure*}[h!]
\centering
\includegraphics[width=2\columnwidth]{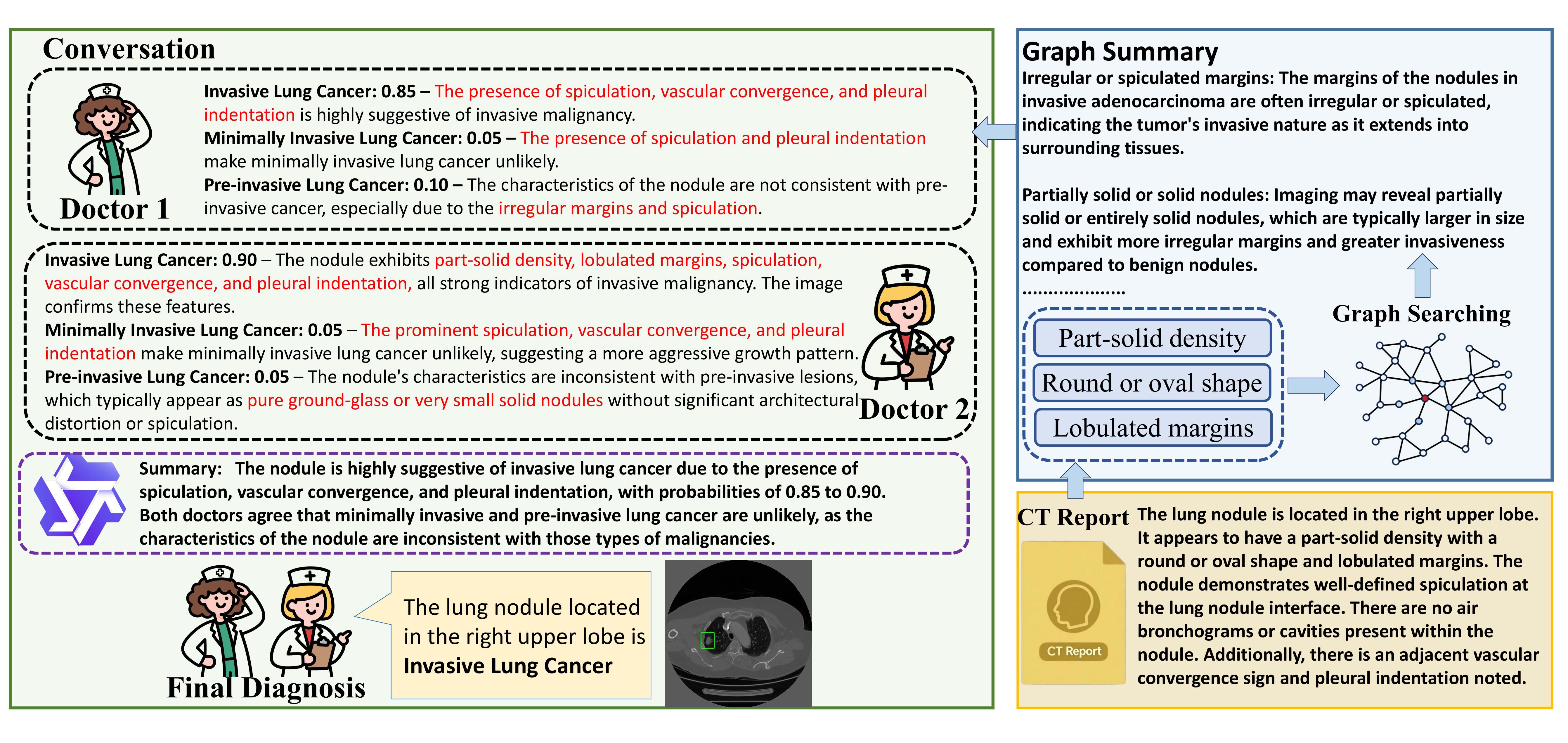}
\caption{Visualization of the DAS's Internal Collaboration. This figure illustrates how multiple agents leverage a medical knowledge graph and a collaborative conversational mechanism to infer and arrive at a final diagnosis for lung nodules based on the CT Report and Nodule Image. }
\label{fig:4}
\end{figure*}

\subsection{Doctor Agent System}
\subsubsection{Medical Graph RAG.}

To address the limited pathological knowledge of existing medical agents, we have adopted a medical graph construction and retrieval strategy inspired by GraphRag \cite{edge2024local}, enabling these agents to access relevant pathological information. This resource aids in constructing reasoning chains and providing diagnostic justifications based on CT reports and medical knowledge. While many medical large language models utilize Retrieval-Augmented Generation (RAG) \cite{arslan2024survey} to obtain external knowledge, they often encounter difficulties with broad queries, such as identifying the primary characteristics of malignant lung nodules, because these questions necessitate query-focused summarization rather than straightforward retrieval.

To address this issue, we propose a Graph Searching framework that compiles data from authoritative pathological websites and literature to create a comprehensive knowledge base. Specifically, the knowledge graph \(\mathcal{G}\) is constructed from documents \(\mathcal{D}\):
\begin{equation}
\mathcal{G} = \text{GraphConstruct}(\mathcal{D}).
\end{equation}

Community-level summaries \(\mathcal{S}\) are then derived from the graph. When presented with a user query \(Q\), the system uses a language model to generate the final answer \(\mathcal{A}\):
\begin{equation}
\mathcal{A} = \text{VLM}(\mathcal{S}, Q, \mathcal{N}),
\end{equation}
where \(\mathcal{N}\) represents the nodule image, defined as the image \(I\) covered with mask \(M\).

Through this graph-based knowledge infusion pipeline (\(\mathcal{D} \rightarrow \mathcal{G} \rightarrow \mathcal{S} \rightarrow \mathcal{A}\)), the large language model evolves into a reasoning agent with specialized medical expertise. This transformation enhances its ability to deliver accurate and detailed diagnostic insights.

\subsubsection{Multi-Agent System.}

Our multi-agent system, illustrated in \textbf{Figure \ref{fig:4}}, employs a decentralized and collaborative approach akin to a roundtable discussion. This methodology is designed to enhance diagnoses, verify information, and ensure the precision of reasoning. At the outset, \(K\) reasoning agents independently assess the report and image \(I\), each generating their initial outputs:
\begin{equation}
O_i^{(1)} = \text{Agent}_i(I, \text{Report}),\quad i = 1,\dots,K,
\end{equation}
where \(i\) denotes the \(i\)th agent. If discrepancies occur, agents refine their conclusions by considering the perspectives of other agents:
\begin{equation}
O_i^{(t)} = \text{Revise}(O_i^{(t-1)}, \{O_j^{(t-1)}\}_{j \ne i}).
\end{equation}

Subsequently, a summarization agent consolidates these outputs into a cohesive summary. This iterative process continues until a consensus is achieved:
\begin{equation}
\mathcal{FD} = \text{Consensus}\left(\text{Summarizer}\left(\{O_i^{(t^*)}\}_{i=1}^K\right)\right),
\end{equation}
where \(\mathcal{FD}\) represents the final diagnosis.

Through this multi-round communication and refinement, the system delivers reliable diagnostic results grounded in collective agreement among the agents.

\subsection{Memory}
The Memory Module serves as the system’s core storage component, designed for managing critical information. It stores nodule images, while measuring and storing nodule size as evidence. It also holds CT Reports, which are initial interpretations of visual image features. During the DAS discussion process, the Memory Module stores detailed Conversations and Summaries from multi-agent interactions. This layered and real-time memory mechanism facilitates a more intuitive understanding of lung nodules by the agents and reduces information processing complexity.

\section{Experiments}
\subsection{Datasets and Implementation}

Our experiments utilized two private datasets comprised of 1,616 and 386 axial lung CT slices (512×512 pixels). These datasets are carefully annotated to aid lung nodule detection, featuring bounding box masks, comprehensive morphological descriptions including lobar location, density, shape, margin, and indications of cavitation or vacuolation, and classifications of malignancy, categorized as pre-invasive, minimally invasive, or invasive adenocarcinoma. In addition to these private sources, we employed the public dataset LIDC-IDRI \cite{samuel2011lung}, including 1,018 lung CT scans with identical resolution. This dataset provides detailed lung nodule segmentation masks and semantic attributes such as size, margin clarity, density, and spiculation. Furthermore, it offers malignancy ratings ranging from 1 to 5, classifying scores above 3 as malignant and those below as benign.

\subsection{Attribute Question Construction}
To address the challenge of lacking precise ground truth for localized CT reports, we adopt an attribute-level verification approach inspired by the MedDLC-score \cite{xiao2025describe}. For each nodule, we develop a series of clinically relevant yes/no questions based on dataset annotations to evaluate if the generated descriptions accurately capture key morphological traits such as shape, margin, and size. These questions are split into two categories: positive questions (Pos QA) verify the presence of specific nodule characteristics, while negative questions (Neg QA) detect irrelevant or fabricated details. The average correctness rate of these questions yields the LungDLC-score, a reliable metric for assessing the accuracy with which reports capture essential clinical features and morphology.

\begin{table*}[t]
\centering
\resizebox{\linewidth}{!}{
\begin{tabular}{c|c|cccc|cccc|cccc}
\toprule
\multicolumn{2}{c|}{\multirow{2}{*}{Method}}
& \multicolumn{4}{c|}{PrivateA} 
& \multicolumn{4}{c|}{PrivateB} 
& \multicolumn{4}{c}{LIDC-IDRI} \\
\cmidrule{3-14}
\multicolumn{2}{c|}{} 
& LLM-score & LungDLC & Pos QA & Neg QA 
& LLM-score & LungDLC & Pos QA & Neg QA 
& LLM-score & LungDLC & Pos QA & Neg QA \\
\midrule
\multirow{5}{*}{\rotatebox{90}{\textbf{Generalist}}} 
& GPT-4o & \textbf{88.1} & 71.8 & 68.3 & 75.3 
       & \underline{87.2} & 68.2 & 66.5 & 69.8 
       & \textbf{90.1} & 73.2 & 70.1 & 76.3 \\
& Claude 3.7 Sonnet & 86.9 & 65.6 & 58.8 & 72.3 
                 & 86.5 & 64.3 & 55.2 & 73.3 
                 & 88.6 & 65.9 & 59.2 & 72.6 \\
& Qwen2.5-VL-7B$^\triangle$& 80.2 & 54.3 & 41.4 & 67.2 
              & 78.2 & 56.3 & 48.2 & 64.5 
              & 82.5 & 60.3 & 50.1 & 70.4 \\
& InternVL3-8B & 77.2 & 56.3 & 43.8 & 68.9 
            & 75.8 & 54.6 & 41.9 & 67.4 
            & 77.5 & 61.8 & 54.9 & 68.7 \\
& LLaVA3.2-11B$^\Box$& 83.1 & 58.1 & 46.1 & 70.2 
      & 81.7 & 60.3 & 52.8 & 67.8 
      & 84.2 & 62.3 & 55.8 & 69.8 \\
\midrule
\multirow{6}{*}{\rotatebox{90}{\textbf{Medical Agent}}} 
& MedGemma-27B & 87.3 & 75.6 & 76.0 & 75.2
         & 87.0 & 76.2 & \underline{73.7} & 78.6
         & 87.7 & 75.2 & 74.0 & 76.4 \\
& MedR1-3B & 82.1 & 64.5 & 55.4 & 73.5 
      & 80.5 & 61.1 & 49.9 & 72.3 
      & 83.3 & 67.2 & 62.7 & 71.7 \\
& MedAgent-Pro & 86.9 & 70.4 & 62.7 & 78.1
      & 85.6 & 68.8 & 63.0 & 74.6
      & 87.2 & 70.0 & 66.8 & 73.2 \\
& MedAgents & 83.6 & 65.2 & 55.7 & 74.7 
      & 82.6 & 63.7 & 54.1 & 73.3 
      & 84.0 & 68.5 & 61.3 & 75.7 \\
& MDAgent & 77.8 & 63.3 & 54.5 & 72.1
      & 81.3 & 60.5 & 48.6 & 72.5
      & 82.6 & 63.2 & 52.8 & 73.7 \\
& LLaVA-Med & 77.6 & 58.3 & 50.1 & 66.5 
      & 80.2 & 58.9 & 47.5 & 70.2
      & 82.1 & 61.5 & 53.3 & 69.6 \\
\midrule
\multirow{4}{*}{\rotatebox{90}{\textbf{Ours}}} 
& \textbf{LungNoduleAgent} & 86.9 & \underline{80.5} & \underline{77.3} & \textbf{83.7} 
            & 86.2 & \underline{79.5} & 73.6 & \textbf{85.4} 
            & 88.5 & \underline{83.1} & \underline{81.6} & \underline{84.6} \\
& Qwen-2.5-VL-7B$^\triangle$& \multicolumn{1}{r}{+6.7} & \multicolumn{1}{r}{+26.2} & \multicolumn{1}{r}{+35.9} & \multicolumn{1}{r|}{+16.5} & \multicolumn{1}{r}{+8.0} & \multicolumn{1}{r}{+23.2} & \multicolumn{1}{r}{+25.4} & \multicolumn{1}{r|}{+20.9} & \multicolumn{1}{r}{+6.0} & \multicolumn{1}{r}{+22.8} & \multicolumn{1}{r}{+31.5} & \multicolumn{1}{r}{+14.2}  \\
& \textbf{LungNoduleAgent} & \underline{87.6} & \textbf{81.9} & \textbf{81.6} & \underline{82.2} 
            & \textbf{89.8} & \textbf{80.3} & \textbf{76.1} & \underline{84.5} 
            & \underline{89.3} & \textbf{83.5} & \textbf{82.1} & \textbf{84.9} \\
& LLaVA3.2-11B$^\Box$& \multicolumn{1}{r}{\textcolor{black}{+4.5}} & \multicolumn{1}{r}{\textcolor{black}{+23.8}} & \multicolumn{1}{r}{\textcolor{black}{+35.5}} & \multicolumn{1}{r|}{\textcolor{black}{+12.0}} & \multicolumn{1}{r}{\textcolor{black}{+8.1}} & \multicolumn{1}{r}{\textcolor{black}{+20.0}} & \multicolumn{1}{r}{\textcolor{black}{+23.3}} & \multicolumn{1}{r|}{\textcolor{black}{+16.7}} & \multicolumn{1}{r}{\textcolor{black}{+5.1}} & \multicolumn{1}{r}{\textcolor{black}{21.2}} & \multicolumn{1}{r}{\textcolor{black}{+26.3}} & \multicolumn{1}{r}{\textcolor{black}{+15.1}} \\
\bottomrule
\end{tabular}}
\caption{CT report generation results on three datasets. The best and second-best results are \textbf{bolded} and \underline{underlined}, respectively. $\triangle, \Box $ correspond to the base models of our LungNoduleAgent for
comparisons, and + shows the improvements between them.}
\label{tab:1}
\end{table*}

\begin{table}[t]
\centering
\resizebox{\linewidth}{!}{
\begin{tabular}{c|l|cc|cc|cc} 
\toprule
\multicolumn{2}{c|}{\multirow{2}{*}{Method}}
& \multicolumn{2}{c|}{PrivateA} 
& \multicolumn{2}{c|}{PrivateB} 
& \multicolumn{2}{c}{LIDC-IDRI} \\
\cmidrule{3-8}
\multicolumn{2}{c|}{} & Acc(\%) & F1($10^{-2}$) & Acc(\%) & F1($10^{-2}$) & Acc(\%) & F1($10^{-2}$) \\
\midrule
\multirow{5}{*}{\rotatebox{90}{\textbf{Generalist}}} 
& GPT-4o & 46.2 & 39.7 & 41.2 & 56.2 & 64.1 & 76.7 \\
& Claude 3.7 Sonnet & 33.1 & 35.6 & 34.2 & 39.8 & 62.3 & 68.7 \\
& Qwen2.5-VL-7B$^\triangle$ & 31.4 & 35.4 & 31.1 & 36.8 & 56.5 & 60.2 \\
& InternVL3-8B & 36.1 & 39.8 & 35.3 & 44.6 & 58.3 & 60.8 \\
& LLaVA3.2-11B$^\Box$ & 35.6 & 40.3 & 36.7 & 45.2 & 59.1 & 61.2 \\
\midrule
\multirow{6}{*}{\rotatebox{90}{\textbf{Medical Agent}}} 
& MedGemma-27B & 62.3 & 68.3 & 60.2 & 70.6 & 73.2 & 68.6 \\
& MedR1-3B & 52.3 & 62.7 & 48.9 & 58.6 & 67.6 & 73.5 \\
& MedAgent-Pro & 60.2 & 65.9 & 60.1 & 71.9 & 72.6 & 74.1 \\
& MedAgents & 54.3 & 64.2 & 52.3 & 60.1 & 65.3 & 58.2 \\
& MDAgent & 45.6 & 52.1 & 43.9 & 56.5 & 66.7 & 63.2 \\
& LLaVA-Med & 39.3 & 38.2 & 40.7 & 45.9 & 62.5 & 65.8 \\
\midrule
\multirow{4}{*}{\rotatebox{90}{\textbf{Ours}}} 
& \textbf{LungNoduleAgent} & \underline{85.8} & \underline{82.1} & \underline{80.6} & \underline{75.6} & \underline{88.5} & \underline{84.1} \\
& Qwen-2.5-VL-7B$^\triangle$& \multicolumn{1}{r}{+54.4} & \multicolumn{1}{r}{+46.7} & \multicolumn{1}{r}{+49.5} & \multicolumn{1}{r|}{+38.8} & \multicolumn{1}{r}{+32.0} & \multicolumn{1}{r}{+23.9} \\
& \textbf{LungNoduleAgent} & \textbf{86.7} & \textbf{88.9} & \textbf{81.2} & \textbf{80.3} & \textbf{89.1} & \textbf{87.1} \\
& LLaVA3.2-11B$^\Box$& \multicolumn{1}{r}{+51.1} & \multicolumn{1}{r}{+48.6} & \multicolumn{1}{r}{+44.5} & \multicolumn{1}{r|}{+35.1} & \multicolumn{1}{r}{+30.0} & \multicolumn{1}{r}{+25.9} \\
\bottomrule
\end{tabular}}
\caption{Malignancy prediction results on three datasets. The best and second-best results are \textbf{bolded} and \underline{underlined}, respectively. $\triangle, \Box $ correspond to the base models of our LungNoduleAgent for
comparisons, and + shows the improvements between them.}
\label{tab:4}
\end{table}

\subsection{Comparison Methods and Evaluation Metrics}

We assessed our proposed method against a selection of current models, including general-purpose VLMs like GPT-4o \cite{achiam2023gpt}, Claude 3.7 Sonnet \cite{anthropic2024claude}, InternVL \cite{zhu2025internvl3}, LLaVA3.2 \cite{liu2023llava}, and Qwen2.5-VL \cite{Qwen2.5-VL}. Additionally, we evaluated medical agents such as Med-R1 \cite{lai2025med}, MedGamma \cite{sellergren2025medgemma}, MedAgent-Pro \cite{wang2025medagent}, MDAgents \cite{kim2024mdagents}, MedAgents \cite{tang2023medagents}, and LLaVA-Med \cite{li2023llavamed}. These models epitomize the latest in visual-language pre-training and domain-specific tuning advancements. To ensure consistency, all baseline models were given full-slice lung CT images and the same MedPrompt.

We evaluate CT report generation using a two-pronged approach. First, we utilize ``LLM-as-a-judge'' framework \cite{li2024llmasajudge}, where GPT-4o rates generated reports for fluency, relevance, consistency, and clinical rationality. Each aspect is scored independently, and their average forms the LLM-score. Second, to reduce reliance on ground truth, we introduce the LungDLC-score, which checks whether descriptions capture nodule features and avoid irrelevant or fabricated content. For malignancy grading, we report Accuracy (Acc) and F1-score.

\subsection{Results and Discussions}

\textbf{Results on localized CT report generation task.}
As demonstrated in \textbf{Table \ref{tab:1}}, LungNoduleAgent exhibits state-of-the-art performance across all three benchmark datasets, with significant improvements on the Qwen and LLaVA base models. Compared with the highest score among other methods, on the PrivateA dataset, it achieves a LungDLC-score of 81.9, reflecting an increase of 6.3. For the PrivateB dataset, the model attains a score of 80.3, marking an improvement of 4.1. This performance advantage is also evident on the public LIDC-IDRI benchmark, where the model scores 83.5, representing 8.3 increasing.

Notably, the agent demonstrates balanced capabilities in both Positive QA and Negative QA . While GPT-4o shows marginally higher LLM-scores due to evaluator bias \cite{li2025preference}, our model's consistent superiority in objective metrics validates its clinical utility.




\textbf{The result of the malignancy grading task.} 
LungNoduleAgent achieved state-of-the-art performance across all evaluation datasets, with significant improvements on both Qwen and LLaVA base models, demonstrating strong generalization capability. On the 3-class classification tasks (PrivateA/PrivateB), it attained Acc of 86.7\% and 81.2\% with corresponding F1-scores of 0.889 and 0.803, outperforming Medgamma by 15.9-24.4\% in Acc. For the 2-class LIDC-IDRI benchmark, our method achieved 89.1\% Acc and 0.871 F1, surpassing MedGemma by 15.9\% in Acc and 0.185 in F1. The consistent superiority across different classification tasks and datasets validates the effectiveness of our nodule-focused architecture and multi-agent reasoning framework for clinical malignancy assessment.



\begin{table}[h!]
\centering
\begin{tabular}{ccc|cc}
\toprule
NS & SR & DAS & Acc(\%) & LungDLC \\
\midrule
- & - & \checkmark & 62.1 & 57.9 \\
\checkmark & \checkmark & - & 66.7 & \underline{88.9} \\
\checkmark & - & \checkmark & \underline{75.1} & 67.3 \\
\checkmark & \checkmark & \checkmark & \textbf{86.7} & \textbf{88.9} \\
\bottomrule
\end{tabular}
\caption{Ablation study on different module combinations. The best and second-best results are \textbf{bolded} and \underline{underlined}, respectively. Abbreviations: Nodule Spotter (NS), Simulated Radiologist (SR), Doctor Agent System (DAS).}
\label{tab:3}
\end{table}

\textbf{Ablation Study.} We performed an ablation study on the PrivateA dataset to assess the effectiveness of each module within our framework: Nodule Spotter, Simulated Radiologist, and DAS. It is noteworthy that the Nodule Spotter cannot be independently ablated, as the Simulated Radiologist relies on the masks it provides. As shown in \textbf{Table \ref{tab:3}}, the results revealed a significant decline in performance across all tasks when any module was removed. This underscores the effectiveness of each component in our framework.

\begin{table}[h!]
\centering
\begin{tabular}{ccc|cc}
\toprule
MoE & Clustering & Judge Panel & mAP(\%) & F1($10^{-2}$) \\
\midrule
\checkmark & - & - & 67.1 & 64.3 \\
\checkmark & \checkmark & - & \underline{71.6} & \underline{71.3} \\
\checkmark & \checkmark & \checkmark & \textbf{79.3} & \textbf{83.5} \\
\bottomrule
\end{tabular}
\caption{Ablation study on the effect of MoE, Clustering, and Judge Panel. The best and second-best results are \textbf{bolded} and \underline{underlined}, respectively.}
\label{tab:2}
\end{table}

We examined the effectiveness of the Mask Clustering and Judge Panel components within the Nodule Spotter module, where each component builds upon the previous one. To assess their impact, we used standard object detection metrics, specifically mean Average Precision (mAP) and F1-score. As shown in \textbf{Table \ref{tab:2}}, integrating Mask Clustering and the Judge Panel led to mAP improvements of 4\% and 8\%, respectively, and F1-score enhancements of 0.07 and 0.12, respectively. These results clearly demonstrate the effectiveness of both components in refining anomalous masks and improving the overall quality of mask generation.

To understand the impact of the Nodule Spotter on the overall framework's performance, we artificially created multiple detection regions with different IoU values compared to ground truth masks. We consistently found that higher detection Acc led to better diagnostic outcomes. As illustrated in \textbf{Figure \ref{fig:3}(a)}, these findings highlight the importance of evidence-based quantitative analysis over experience-driven qualitative assessments in multimodal diagnostic contexts, thereby affirming the effectiveness of our proposed framework.

We conducted an ablation study to assess how varying the number of medical agents influences the DAS's Acc in grading malignancies, as illustrated in \textbf{Figure \ref{fig:3}(b)}. The findings show that Acc increased with additional agents, peaking at five agents, after which the performance became inconsistent \cite{xiong2023examining}. This trend suggests that a broader range of perspectives enhances diagnostic reasoning up to an optimal point. Beyond this threshold, the inclusion of more agents may introduce redundancy or minor inconsistencies, leading to fluctuations in performance. Consequently, we determined that deploying five medical agents strikes the ideal balance between robust diagnostic performance and computational efficiency in our system's analyses.

\begin{figure}[h!]
    \centering
    \begin{tabular}{cc}
        \includegraphics[width=0.22\textwidth]{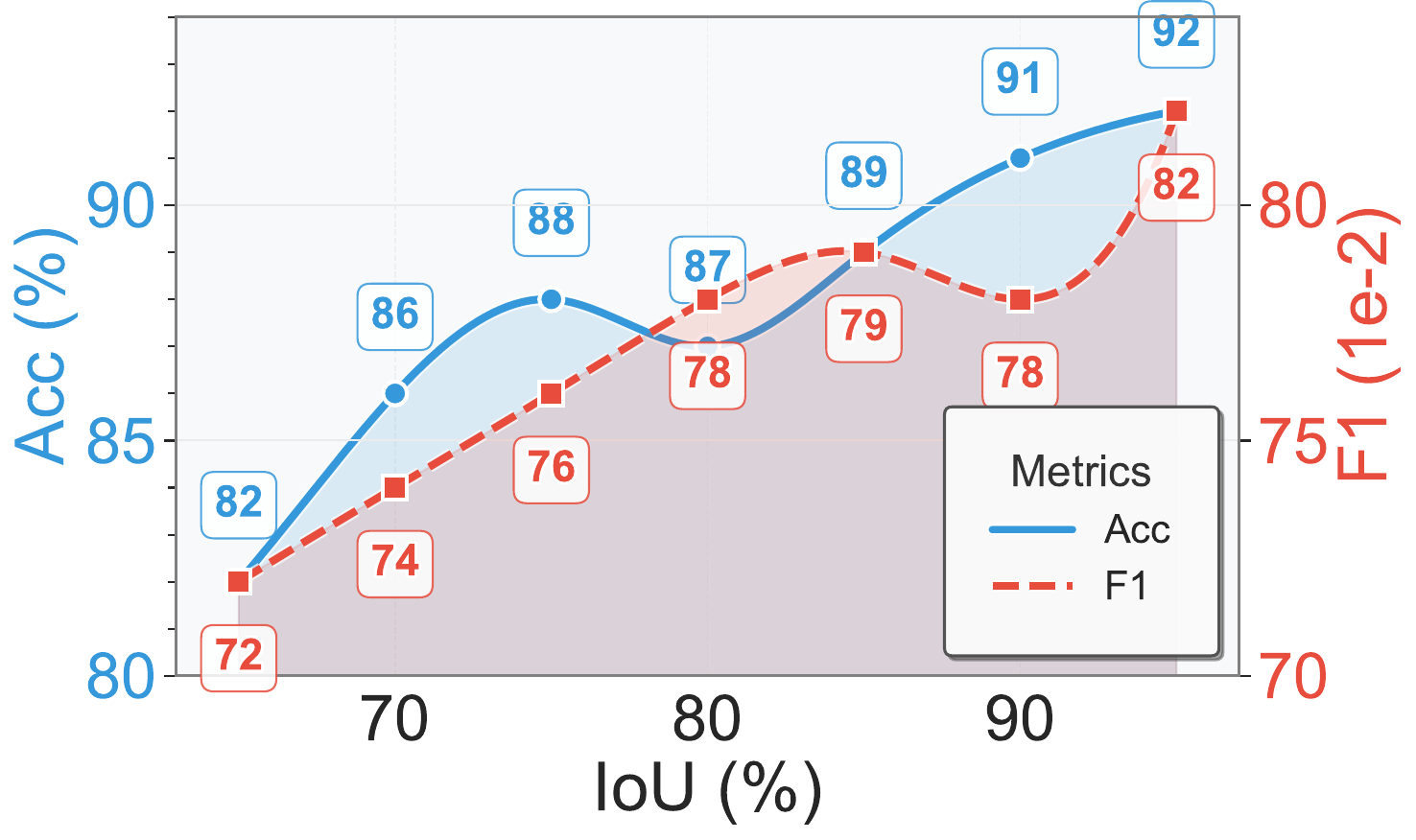} &
        \includegraphics[width=0.22\textwidth]{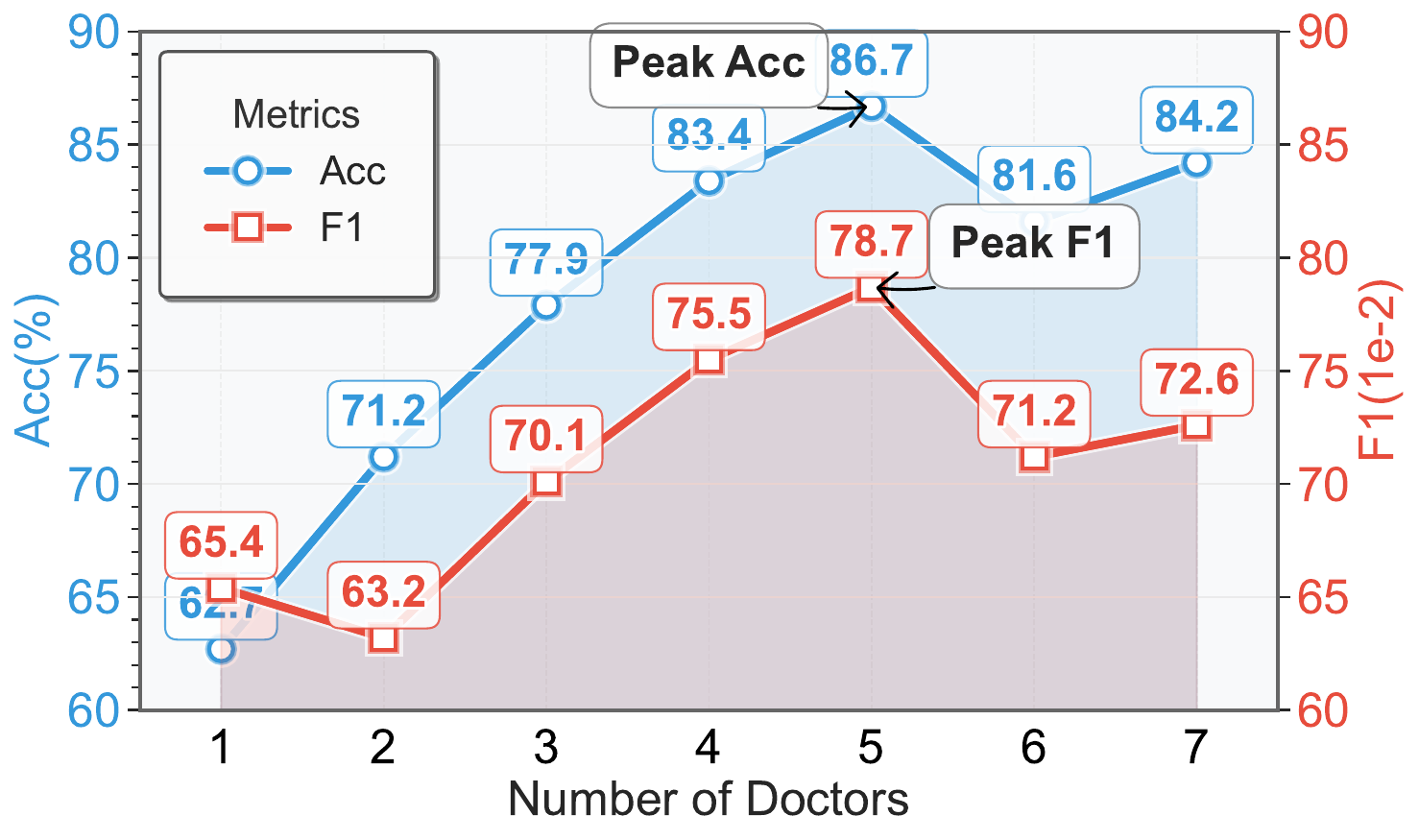} \\
        (a) & (b) \\
    \end{tabular}
    \caption{(a): Ablation study on the influence of nodule detection Acc on malignancy grading. (b): Ablation study on the impact of the number of medical agents in the DAS on nodule malignancy grading.}
    \label{fig:3}
\end{figure}

\section{Conclusion}

We introduce LungNoduleAgent, a novel multi-agent system designed to mimic the clinical workflow of lung nodule diagnosis in CT scans. Our framework is specifically tailored to produce accurate malignancy grading and detailed CT reports, proving the significance of region-level semantic alignment and multi-agent collaboration in diagnosing nodules. Compared to existing VLMs and medical agents, our framework shows superior performance and remarkable generalization capabilities. Ablation studies further validate the effectiveness of each component within the system.


\section{Acknowledgements}
This work was supported by National Natural Science Foundation of China (No.62076084,  61702146, U20A20386, U22A2033) and Zhejiang Provincial Natural Science Foundation of China (No. LY21F020017, 2023C03090), Guangxi Science and Technology Program (No. FN2504240022), the Guangxi Key R\&D Project (No. AB24010167), the Project (No. 20232ABC03A25), GuangDong Basic and Applied Basic Research Foundation (No. 2025A1515011617, 2022A1515110570), Futian Healthcare Research Project (No.FTWS002), Central Funds Guiding the Local Science and Technology Development Project (No. 2025ZYDF106).

\bibliography{main}

@article{swanson2023patterns,
  title={From patterns to patients: Advances in clinical machine learning for cancer diagnosis, prognosis, and treatment},
  author={Swanson, Kyle and Wu, Eric and Zhang, Angela and Alizadeh, Ash A and Zou, James},
  journal={Cell},
  volume={186},
  number={8},
  pages={1772--1791},
  year={2023},
  publisher={Elsevier}
}

@article{driessen2025completeness,
  title={Completeness and Accuracy of Malignancy History in Abdominal CT Order Requisitions and Final Radiology Reports},
  author={Driessen, Rebecca and Nandwana, Sadhna and Hajibonabi, Farid and Moreno, Courtney and Davarpanah, Amir and Balthazar, Patricia},
  journal={Current Problems in Diagnostic Radiology},
  year={2025},
  publisher={Elsevier}
}

@article{thirunavukarasu2023large,
  title={Large language models in medicine},
  author={Thirunavukarasu, Arun James and Ting, Darren Shu Jeng and Elangovan, Kabilan and Gutierrez, Laura and Tan, Ting Fang and Ting, Daniel Shu Wei},
  journal={Nature Mdicine},
  volume={29},
  number={8},
  pages={1930--1940},
  year={2023},
  publisher={Nature Publishing Group US New York}
}

@article{zhang2024mm,
  title={Mm-llms: Recent advances in multimodal large language models},
  author={Zhang, Duzhen and Yu, Yahan and Dong, Jiahua and Li, Chenxing and Su, Dan and Chu, Chenhui and Yu, Dong},
  journal={arXiv preprint arXiv:2401.13601},
  year={2024}
}

@article{lai2025med,
  title={Med-r1: Reinforcement learning for generalizable medical reasoning in vision-language models},
  author={Lai, Yuxiang and Zhong, Jike and Li, Ming and Zhao, Shitian and Yang, Xiaofeng},
  journal={arXiv preprint arXiv:2503.13939},
  year={2025}
}

@article{liu2023llava,
  title={Visual instruction tuning},
  author={Liu, Haotian and Li, Chunyuan and Wu, Qingyang and Lee, Yong Jae},
  journal={Advances in Neural Information Processing Systems},
  volume={36},
  pages={34892--34916},
  year={2023}
}

@article{achiam2023gpt,
  title={Gpt-4 technical report},
  author={Achiam, Josh and Adler, Steven and Agarwal, Sandhini and Ahmad, Lama and Akkaya, Ilge and Aleman, Florencia Leoni and Almeida, Diogo and Altenschmidt, Janko and Altman, Sam and Anadkat, Shyamal and others},
  journal={arXiv preprint arXiv:2303.08774},
  year={2023}
}

@article{tammemagi2019predicting,
  title={Predicting malignancy risk of screen-detected lung nodules--mean diameter or volume},
  author={Tammemagi, Martin and Ritchie, Alex J and Atkar-Khattra, Sukhinder and Dougherty, Brendan and Sanghera, Calvin and Mayo, John R and Yuan, Ren and Manos, Daria and McWilliams, Annette M and Schmidt, Heidi and others},
  journal={Journal of Thoracic Oncology},
  volume={14},
  number={2},
  pages={203--211},
  year={2019},
  publisher={Elsevier}
}

@article{osarogiagbon2023evaluation,
  title={Evaluation of lung cancer risk among persons undergoing screening or guideline-concordant monitoring of lung nodules in the Mississippi Delta},
  author={Osarogiagbon, Raymond U and Liao, Wei and Faris, Nicholas R and Fehnel, Carrie and Goss, Jordan and Shepherd, Catherine J and Qureshi, Talat and Matthews, Anberitha T and Smeltzer, Matthew P and Pinsky, Paul F},
  journal={JAMA Network Open},
  volume={6},
  number={2},
  pages={e230787--e230787},
  year={2023},
  publisher={American Medical Association}
}

@article{arslan2024survey,
  title={A Survey on RAG with LLMs},
  author={Arslan, Muhammad and Ghanem, Hussam and Munawar, Saba and Cruz, Christophe},
  journal={Procedia Computer Science},
  volume={246},
  pages={3781--3790},
  year={2024},
  publisher={Elsevier}
}

@article{zhang2021dual,
  title={Dual attention-in-attention model for joint rain streak and raindrop removal},
  author={Zhang, Kaihao and Li, Dongxu and Luo, Wenhan and Ren, Wenqi},
  journal={IEEE Transactions on Image Processing},
  volume={30},
  pages={7608--7619},
  year={2021},
  publisher={IEEE}
}

@article{sellergren2025medgemma,
  title={MedGemma Technical Report},
  author={Sellergren, Andrew and Kazemzadeh, Sahar and Jaroensri, Tiam and Kiraly, Atilla and Traverse, Madeleine and Kohlberger, Timo and Xu, Shawn and Jamil, Fayaz and Hughes, C{\'\i}an and Lau, Charles and others},
  journal={arXiv preprint arXiv:2507.05201},
  year={2025}
}

@misc{anthropic2024claude,
  author = {Anthropic},
  title = {Claude 3.7 Sonnet},
  year = {2024},
  url = {https://www.anthropic.com/news/claude-3-7-sonnet}
}

@article{xiao2025describe,
  title={Describe Anything in Medical Images},
  author={Xiao, Xi and Zhang, Yunbei and Nguyen, Thanh-Huy and Lam, Ba-Thinh and Wang, Janet and Zhao, Lin and Hamm, Jihun and Wang, Tianyang and Li, Xingjian and Wang, Xiao and others},
  journal={arXiv preprint arXiv:2505.05804},
  year={2025}
}

@article{samuel2011lung,
  title={The Lung Image Database Consortium (LIDC) and Image Database resource initiative (IDRI): A completed reference database of lung nodules on CT scans},
  author={Samuel, G},
  journal={Medical Physics},
  volume={38},
  pages={2},
  year={2011}
}

@article{edge2024local,
  title={From local to global: A graph rag approach to query-focused summarization},
  author={Edge, Darren and Trinh, Ha and Cheng, Newman and Bradley, Joshua and Chao, Alex and Mody, Apurva and Truitt, Steven and Metropolitansky, Dasha and Ness, Robert Osazuwa and Larson, Jonathan},
  journal={arXiv preprint arXiv:2404.16130},
  year={2024}
}

@article{Qwen2.5-VL,
  title={Qwen2.5-VL Technical Report},
  author={Bai, Shuai and Chen, Keqin and Liu, Xuejing and Wang, Jialin and Ge, Wenbin and Song, Sibo and Dang, Kai and Wang, Peng and Wang, Shijie and Tang, Jun and Zhong, Humen and Zhu, Yuanzhi and Yang, Mingkun and Li, Zhaohai and Wan, Jianqiang and Wang, Pengfei and Ding, Wei and Fu, Zheren and Xu, Yiheng and Ye, Jiabo and Zhang, Xi and Xie, Tianbao and Cheng, Zesen and Zhang, Hang and Yang, Zhibo and Xu, Haiyang and Lin, Junyang},
  journal={arXiv preprint arXiv:2502.13923},
  year={2025}
}

@article{zhu2025internvl3,
  title={Internvl3: Exploring advanced training and test-time recipes for open-source multimodal models},
  author={Zhu, Jinguo and Wang, Weiyun and Chen, Zhe and Liu, Zhaoyang and Ye, Shenglong and Gu, Lixin and Tian, Hao and Duan, Yuchen and Su, Weijie and Shao, Jie and others},
  journal={arXiv preprint arXiv:2504.10479},
  year={2025}
}

@article{kim2024mdagents,
  title={Mdagents: An adaptive collaboration of llms for medical decision-making},
  author={Kim, Yubin and Park, Chanwoo and Jeong, Hyewon and Chan, Yik S and Xu, Xuhai and McDuff, Daniel and Lee, Hyeonhoon and Ghassemi, Marzyeh and Breazeal, Cynthia and Park, Hae W},
  journal={Advances in Neural Information Processing Systems},
  volume={37},
  pages={79410--79452},
  year={2024}
}

@article{li2024mmedagent,
  title={Mmedagent: Learning to use medical tools with multi-modal agent},
  author={Li, Binxu and Yan, Tiankai and Pan, Yuanting and Luo, Jie and Ji, Ruiyang and Ding, Jiayuan and Xu, Zhe and Liu, Shilong and Dong, Haoyu and Lin, Zihao and others},
  journal={arXiv preprint arXiv:2407.02483},
  year={2024}
}

@article{li2024llmasajudge,
      title   = {From Generation to Judgment: Opportunities and Challenges of LLM-as-a-judge},
      author  = {Dawei Li and Bohan Jiang and Liangjie Huang and Alimohammad Beigi and Chengshuai Zhao and Zhen Tan and Amrita Bhattacharjee and Yuxuan Jiang and Canyu Chen and Tianhao Wu and Kai Shu and Lu Cheng and Huan Liu},
      year    = {2024},
      journal = {arXiv preprint arXiv: 2411.16594}
}

@article{li2025preference,
  title={Preference leakage: A contamination problem in llm-as-a-judge},
  author={Li, Dawei and Sun, Renliang and Huang, Yue and Zhong, Ming and Jiang, Bohan and Han, Jiawei and Zhang, Xiangliang and Wang, Wei and Liu, Huan},
  journal={arXiv preprint arXiv:2502.01534},
  year={2025}
}

@article{qiu2025embodied,
  title={Embodied artificial intelligence in ophthalmology},
  author={Qiu, Yao and Chen, Xiaolan and Wu, Xinyuan and Li, Yunqian and Xu, Pusheng and Jin, Kai and Shang, Xianwen and Chotcomwongse, Peranut and He, Mingguang and Shi, Danli},
  journal={NPJ Digital Medicine},
  volume={8},
  number={1},
  pages={351},
  year={2025},
  publisher={Nature Publishing Group UK London}
}

@article{ladbury2023integration,
  author={Chad Ladbury and Arya Amini and Anand Govindarajan and others},
  title={Integration of artificial intelligence in lung cancer: Rise of the machine},
  journal={Cell Reports Medicine},
  volume={4},
  number={2},
  year={2023},
}

@article{hammer2019decision,
  title={A decision analysis of follow-up and treatment algorithms for nonsolid pulmonary nodules},
  author={Hammer, Mark M and Palazzo, Lauren L and Eckel, Andrew L and Barbosa Jr, Eduardo M and Kong, Chung Yin},
  journal={Radiology},
  volume={290},
  number={2},
  pages={506--513},
  year={2019},
  publisher={Radiological Society of North America}
}

@article{cao2020two,
  title={A two-stage convolutional neural networks for lung nodule detection},
  author={Cao, Haichao and Liu, Hong and Song, Enmin and Ma, Guangzhi and Xu, Xiangyang and Jin, Renchao and Liu, Tengying and Hung, Chih-Cheng},
  journal={IEEE Journal of Biomedical and Health Informatics},
  volume={24},
  number={7},
  pages={2006--2015},
  year={2020},
  publisher={IEEE}
}

@article{ji2023lung,
  title={Lung nodule detection in medical images based on improved YOLOv5s},
  author={Ji, Zhanlin and Wu, Yun and Zeng, Xinyi and An, Yongli and Zhao, Li and Wang, Zhiwu and Ganchev, Ivan},
  journal={IEEE Access},
  volume={11},
  pages={76371--76387},
  year={2023},
  publisher={IEEE}
}

@article{urrehman2024effective,
  title={Effective lung nodule detection using deep CNN with dual attention mechanisms},
  author={UrRehman, Zia and Qiang, Yan and Wang, Long and Shi, Yiwei and Yang, Qianqian and Khattak, Saeed Ullah and Aftab, Rukhma and Zhao, Juanjuan},
  journal={Scientific Reports},
  volume={14},
  number={1},
  pages={3934},
  year={2024},
  publisher={Nature Publishing Group UK London}
}

@article{nakach2024comprehensive,
  title={A comprehensive investigation of multimodal deep learning fusion strategies for breast cancer classification},
  author={Nakach, Fatima-Zahrae and Idri, Ali and Goceri, Evgin},
  journal={Artificial Intelligence Review},
  volume={57},
  number={12},
  pages={327},
  year={2024},
  publisher={Springer}
}

@article{raza2023lung,
  title={Lung-EffNet: Lung cancer classification using EfficientNet from CT-scan images},
  author={Raza, Rehan and Zulfiqar, Fatima and Khan, Muhammad Owais and Arif, Muhammad and Alvi, Atif and Iftikhar, Muhammad Aksam and Alam, Tanvir},
  journal={Engineering Applications of Artificial Intelligence},
  volume={126},
  pages={106902},
  year={2023},
  publisher={Elsevier}
}

@article{martinez2021deep,
  title={Deep residual transfer learning for automatic diagnosis and grading of diabetic retinopathy},
  author={Martinez-Murcia, Francisco J and Ortiz, Andr{\'e}s and Ram{\'\i}rez, Javier and G{\'o}rriz, Juan M and Cruz, Ricardo},
  journal={Neurocomputing},
  volume={452},
  pages={424--434},
  year={2021},
  publisher={Elsevier}
}

@article{li2023llavamed,
  title={Llava-med: Training a large language-and-vision assistant for biomedicine in one day},
  author={Li, Chunyuan and Wong, Cliff and Zhang, Sheng and Usuyama, Naoto and Liu, Haotian and Yang, Jianwei and Naumann, Tristan and Poon, Hoifung and Gao, Jianfeng},
  journal={arXiv preprint arXiv:2306.00890},
  year={2023}
}

@article{tang2023medagents,
  title={Medagents: Large language models as collaborators for zero-shot medical reasoning},
  author={Tang, Xiangru and Zou, Anni and Zhang, Zhuosheng and Li, Ziming and Zhao, Yilun and Zhang, Xingyao and Cohan, Arman and Gerstein, Mark},
  journal={arXiv preprint arXiv:2311.10537},
  year={2023}
}

@article{wang2025medagent,
  title={MedAgent-Pro: Towards Evidence-Based Multi-Modal Medical Diagnosis via Reasoning Agentic Workflow},
  author={Wang, Ziyue and Wu, Junde and Cai, Linghan and Low, Chang Han and Yang, Xihong and Li, Qiaxuan and Jin, Yueming},
  journal={arXiv preprint arXiv:2503.18968},
  year={2025}
}

@article{zhang2023pmc,
  title={Pmc-vqa: Visual instruction tuning for medical visual question answering},
  author={Zhang, Xiaoman and Wu, Chaoyi and Zhao, Ziheng and Lin, Weixiong and Zhang, Ya and Wang, Yanfeng and Xie, Weidi},
  journal={arXiv preprint arXiv:2305.10415},
  year={2023}
}

@article{lee2024read,
  title={Read like a radiologist: Efficient vision-language model for 3d medical imaging interpretation},
  author={Lee, Changsun and Park, Sangjoon and Shin, Cheong-Il and Choi, Woo Hee and Park, Hyun Jeong and Lee, Jeong Eun and Ye, Jong Chul},
  journal={arXiv preprint arXiv:2412.13558},
  year={2024}
}

@article{liu2023agentbench,
  title={Agentbench: Evaluating llms as agents},
  author={Liu, Xiao and Yu, Hao and Zhang, Hanchen and Xu, Yifan and Lei, Xuanyu and Lai, Hanyu and Gu, Yu and Ding, Hangliang and Men, Kaiwen and Yang, Kejuan and others},
  journal={arXiv preprint arXiv:2308.03688},
  year={2023}
}

@article{liang2023encouraging,
  title={Encouraging divergent thinking in large language models through multi-agent debate},
  author={Liang, Tian and He, Zhiwei and Jiao, Wenxiang and Wang, Xing and Wang, Yan and Wang, Rui and Yang, Yujiu and Shi, Shuming and Tu, Zhaopeng},
  journal={arXiv preprint arXiv:2305.19118},
  year={2023}
}

@inproceedings{du2023improving,
  title={Improving factuality and reasoning in language models through multiagent debate},
  author={Du, Yilun and Li, Shuang and Torralba, Antonio and Tenenbaum, Joshua B and Mordatch, Igor},
  booktitle={Forty-first International Conference on Machine Learning},
  pages={11733--11763},
  year={2023}
}

@article{chan2023chateval,
  title={Chateval: Towards better llm-based evaluators through multi-agent debate},
  author={Chan, Chi-Min and Chen, Weize and Su, Yusheng and Yu, Jianxuan and Xue, Wei and Zhang, Shanghang and Fu, Jie and Liu, Zhiyuan},
  journal={arXiv preprint arXiv:2308.07201},
  year={2023}
}

@article{wang2024beyond,
  title={Beyond direct diagnosis: LLM-based multi-specialist agent consultation for automatic diagnosis},
  author={Wang, Haochun and Zhao, Sendong and Qiang, Zewen and Xi, Nuwa and Qin, Bing and Liu, Ting},
  journal={arXiv preprint arXiv:2401.16107},
  year={2024}
}

@article{liu2024medchain,
  title={Medchain: Bridging the gap between llm agents and clinical practice through interactive sequential benchmarking},
  author={Liu, Jie and Wang, Wenxuan and Ma, Zizhan and Huang, Guolin and SU, Yihang and Chang, Kao-Jung and Chen, Wenting and Li, Haoliang and Shen, Linlin and Lyu, Michael},
  journal={arXiv preprint arXiv:2412.01605},
  year={2024}
}

@article{xiong2023examining,
  title={Examining inter-consistency of large language models collaboration: An in-depth analysis via debate},
  author={Xiong, Kai and Ding, Xiao and Cao, Yixin and Liu, Ting and Qin, Bing},
  journal={arXiv preprint arXiv:2305.11595},
  year={2023}
}

@article{fallahpour2025medrax,
  title={Medrax: Medical reasoning agent for chest x-ray},
  author={Fallahpour, Adibvafa and Ma, Jun and Munim, Alif and Lyu, Hongwei and Wang, Bo},
  journal={arXiv preprint arXiv:2502.02673},
  year={2025}
}

@article{mao2025ct,
  title={CT-Agent: A Multimodal-LLM Agent for 3D CT Radiology Question Answering},
  author={Mao, Yuren and Xu, Wenyi and Qin, Yuyang and Gao, Yunjun},
  journal={arXiv preprint arXiv:2505.16229},
  year={2025}
}

@inproceedings{fan2025deep,
  title={Deep Neural Network for Lung Adenocarcinoma Subtype from Multimodal Fusion of Imaging and Clinical Data},
  author={Fan, Chenchen and Liu, Liya and Wang, Yuxuan and Li, Danna and Liang, Qinghua and Elazab, Ahmed and Liu, Zhou and Hu, Jiahang and Tian, Yuan and Zhang, Yongquan and others},
  booktitle={2025 IEEE 22nd International Symposium on Biomedical Imaging (ISBI)},
  pages={1--5},
  year={2025},
  organization={IEEE}
}

@inproceedings{shen2025csf,
  title={CSF-NET: Cross-Modal Spatiotemporal Fusion Network for Pulmonary Nodule Malignancy Predicting},
  author={Shen, Yin and Fang, Zhaojie and Zhuang, Ke and Zhou, Guanyu and Yu, Xiao and Zhao, Yucheng and Tian, Yuan and Ge, Ruiquan and Wang, Changmiao and Fan, Xiaopeng and others},
  booktitle={2025 IEEE 22nd International Symposium on Biomedical Imaging (ISBI)},
  pages={1--5},
  year={2025},
  organization={IEEE}
}

@article{wang2018automated,
  title={Automated chest screening based on a hybrid model of transfer learning and convolutional sparse denoising autoencoder},
  author={Wang, Changmiao and Elazab, Ahmed and Jia, Fucang and Wu, Jianhuang and Hu, Qingmao},
  journal={Biomedical Engineering Online},
  volume={17},
  number={1},
  pages={63},
  year={2018},
  publisher={Springer}
}

@article{wang2017lung,
  title={Lung nodule classification using deep feature fusion in chest radiography},
  author={Wang, Changmiao and Elazab, Ahmed and Wu, Jianhuang and Hu, Qingmao},
  journal={Computerized Medical Imaging and Graphics},
  volume={57},
  pages={10--18},
  year={2017},
  publisher={Elsevier}
}

@incollection{koonce2021efficientnet,
  title={EfficientNet},
  author={Koonce, Brett},
  booktitle={Convolutional neural networks with swift for Tensorflow: image recognition and dataset categorization},
  pages={109--123},
  year={2021},
  publisher={Springer}
}

@misc{Ultralytics2020,
author = {Ultralytics},
title = {{YOLOv5}},
year = {2020},
howpublished = {\url{https://github.com/ultralytics/yolov5}}
}

@article{zhang2023large,
  title={Large-scale domain-specific pretraining for biomedical vision-language processing},
  author={Zhang, Sheng and Xu, Yanbo and Usuyama, Naoto and Bagga, Jaspreet and Tinn, Robert and Preston, Sam and Rao, Rajesh and Wei, Mu and Valluri, Naveen and Wong, Cliff and others},
  journal={arXiv preprint arXiv:2303.00915},
  volume={2},
  number={3},
  pages={6},
  year={2023}
}

@inproceedings{hu2024omnimedvqa,
  title={Omnimedvqa: A new large-scale comprehensive evaluation benchmark for medical lvlm},
  author={Hu, Yutao and Li, Tianbin and Lu, Quanfeng and Shao, Wenqi and He, Junjun and Qiao, Yu and Luo, Ping},
  booktitle={Proceedings of the IEEE/CVF Conference on Computer Vision and Pattern Recognition},
  pages={22170--22183},
  year={2024}
}

@inproceedings{yu2025small,
  title={Small Lesions-aware Bidirectional Multimodal Multiscale Fusion Network for Lung Disease Classification},
  author={Yu, Jianxun and Ge, Ruiquan and Wang, Zhipeng and Yang, Cheng and Lin, Chenyu and Fu, Xianjun and Liu, Jikui and Elazab, Ahmed and Wang, Changmiao},
  booktitle={International Conference on Medical Image Computing and Computer-Assisted Intervention},
  pages={589--598},
  year={2025},
  organization={Springer}
}

@article{yang2025multi,
  title={From What to Why: A Multi-Agent System for Evidence-based Chemical Reaction Condition Reasoning},
  author={Yang, Cheng and Lu, Jiaxuan and Wan, Haiyuan and Yu, Junchi and Qin, Feiwei},
  journal={arXiv preprint arXiv:2509.23768},
  year={2025}
}

@article{yu2025visual,
  title={Visual document understanding and question answering: A multi-agent collaboration framework with test-time scaling},
  author={Yu, Xinlei and Chen, Zhangquan and Zhang, Yudong and Lu, Shilin and Shen, Ruolin and Zhang, Jiangning and Hu, Xiaobin and Fu, Yanwei and Yan, Shuicheng},
  journal={arXiv preprint arXiv:2508.03404},
  year={2025}
}

@article{yu2025snow,
  title={Visual Multi-Agent System: Mitigating Hallucination Snowballing via Visual Flow},
  author={Yu, Xinlei and Xu, Chengming and Zhang, Guibin and He, Yongbo and Chen, Zhangquan and Xue, Zhucun and Zhang, Jiangning and Liao, Yue and Hu, Xiaobin and Jiang, Yu-Gang and others},
  journal={arXiv preprint arXiv:2509.21789},
  year={2025}
}

\appendix

\clearpage

\begin{center}
\Large\textbf{Supplementary Material for Paper Titled \\``LungNoduleAgent: \\A Collaborative Multi-Agent System \\ for Precision Diagnosis of Lung Nodules"}
\end{center}
\section{Related Work}
\subsection{Deep Learning Method}
Recent advancements in deep learning have significantly transformed lung cancer analysis using CT imaging, focusing on three crucial areas: classification, grading, and nodule detection. In lung cancer classification, convolutional neural networks (CNNs) have excelled. Nakach et al. \cite{nakach2024comprehensive} introduced a multimodal framework that develops specialized models for different input modalities, combining their features for final decision-making. Raza et al. \cite{raza2023lung} tackled the class imbalance issue through various data augmentation techniques and proposed an EfficientNet-based variant \cite{koonce2021efficientnet} for high-performance lung cancer diagnosis. Similarly, Ji et al. \cite{ji2023lung} presented a new YOLOv5s-based framework \cite{Ultralytics2020} for simultaneous lung nodule detection and classification through a two-stage training process.

For tumor grading, deep learning methods have shown promising potential in evaluating cancer aggressiveness. Fan et al. \cite{fan2025deep} developed a multi-channel network to analyze tumor heterogeneity using multiple modalities, linking imaging features with clinical text for grading. Shen et al. \cite{shen2025csf} devised a dynamic fusion mechanism that integrates text and image data, while Martinez et al. \cite{martinez2021deep} combined deep residual CNNs with transfer learning. Wang et al.'s \cite{wang2018automated,wang2017lung} automated system introduced radiomics-guided deep learning, employing an encoding-decoding approach.

In pulmonary nodule detection, recent studies have tackled the challenges of detecting small objects in 3D volumes. Cao et al. \cite{cao2020two} proposed a two-stage framework featuring a candidate generation network followed by a false positive reduction module, enhancing sensitivity at one false positive per scan. Urrehman et al. \cite{urrehman2024effective} developed a detection system using center point prediction, incorporating dual attention mechanisms.

Despite these technical advancements, substantial challenges persist in clinical deployment. The opaque nature of deep learning models, especially complex architectures like transformers and 3D CNNs, poses interpretability issues. While some strategies, such as attention visualization \cite{raza2023lung} and feature importance analysis \cite{nakach2024comprehensive}, strive to improve interpretability, their clinical relevance remains limited. Another critical challenge is dataset bias, as most models are trained on curated datasets that may not represent real-world clinical diversity. Additionally, the task-specific design of many systems restricts their practical application, as radiologists typically require comprehensive solutions that integrate detection, characterization, and follow-up assessment. Recent initiatives to create unified frameworks \cite{swanson2023patterns} and incorporate clinical knowledge \cite{shen2025csf} show promise in advancing towards more clinically viable systems.

\subsection{General VLMs}
Recent advances in general-purpose Vision-Language Models (VLMs) have exhibited impressive capabilities in multimodal understanding. For example, GPT-4o \cite{achiam2023gpt} utilizes a unified transformer architecture that processes visual inputs via a pixel-to-semantic tokenizer. Similarly, the LLaVA framework \cite{liu2023llava} introduces a projection module that aligns CLIP’s visual features with language model embeddings, allowing for zero-shot reasoning on new image-text tasks. Furthering this approach, Qwen2.5VL \cite{Qwen2.5-VL} employs a dynamic token resampling mechanism to adaptively prioritize significant image regions, while InternVL \cite{zhu2025internvl3} achieves state-of-the-art performance with a 6-billion parameter cross-modal encoder trained on 1.2 billion web-crawled image-text pairs.

Despite these advances, three fundamental limitations pose challenges in medical applications. Firstly, the pretraining datasets for these models primarily consist of natural images, creating a domain gap when processing medical imaging modalities. For example, Claude 3.7 Sonnet \cite{anthropic2024claude} exhibits considerably lower accuracy in generating radiology reports compared to captioning natural images. Secondly, the lack of integration of medical domain knowledge results in clinically implausible outputs, with models often producing incorrect anatomical references in medical image interpretations. Thirdly, the coarse-grained visual processing in these models is inadequate for medical imaging specifics, leading to notably lower performance on specialized diagnostic tasks compared to domain-specific models due to insufficient resolution for detecting subtle pathological findings.

\subsection{Medical VLMs}
Targeted medical VLMs aim to overcome the limitations of general-purpose VLMs through specialized architectures and training approaches. For instance, Med-R1 \cite{lai2025med} employs a hybrid architecture that uses cross-modal mechanisms to achieve fine-grained alignment between images and reports, as demonstrated on the OmniMedVQA benchmark \cite{hu2024omnimedvqa}. Similarly, LLaVA-Med \cite{li2023llavamed} adapts the base LLaVA architecture by further pretraining on an extensive dataset of medical image-text pairs from PMC-15M \cite{zhang2023large}, leading to significant improvements in pathology localization tasks.

However, these specialized models still encounter two major challenges. First, their visual perception limitations become apparent in quantitative tasks. For example, MedGemma \cite{sellergren2025medgemma} demonstrates lower precision in measuring tumor size compared to dedicated segmentation models. Second, current medical VLMs tend to rely heavily on parametric knowledge; their performance drops substantially when faced with novel rare diseases not included in the training data, as observed with PMC-15M. This underscores the need for frameworks that incorporate evidence-based reasoning. Recent research \cite{thirunavukarasu2023large} proposes hybrid architectures that integrate medical knowledge graphs, showing promise in combining learned representations with structured clinical ontologies.

\subsection{Multi-Agent System}
Recent advancements in agentic systems have introduced innovative paradigms for medical decision-making through collaborative artificial intelligence frameworks. The AgentBench ecosystem \cite{liu2023agentbench} provides a standardized evaluation platform where multi-agent systems exhibit clinically relevant performance on general clinical reasoning tasks by adopting specialized roles, such as radiologist and pathologist agents, which engage in structured dialogues \cite{liang2023encouraging}. Building on this, ChatEval \cite{chan2023chateval} implements a dynamic debate mechanism where diagnostic disagreements between agents prompt evidence-based verification processes, enhancing diagnostic consistency in mammography cases.

Three primary architectural strategies have emerged in medical agent systems. First, Tool-Augmented Agents like MedAgent \cite{wang2025medagent} incorporate specialized deep learning modules as executable tools, such as a nnUNet-based nodule segmenter and a ResNet-50 malignancy classifier, integrating these tools with GPT-4's reasoning for thoracic imaging analysis. Second, Knowledge-Grounded Systems like MMAgent \cite{li2024mmedagent} employ a retrieval-augmented generation pipeline that queries clinical guidelines from UpToDate during decision-making, significantly reducing hallucination rates in differential diagnosis. Third, Multimodal Workflows like MedRax \cite{fallahpour2025medrax} introduce a radiology-specific agent framework that sequentially processes DICOM metadata, image features, and clinical notes for cross-modal anomaly detection.

Despite these innovations, significant limitations remain in pulmonary applications. Current systems show room for improvement in lung cancer staging tasks \cite{tang2023medagents}, primarily due to three factors: challenges in addressing nodule heterogeneity (solid, ground-glass, part-solid subtypes), where MedChain \cite{liu2024medchain} struggles with mixed-density nodules; limited incorporation of temporal data, with longitudinal analysis providing more modest benefits compared to human radiologists \cite{kim2024mdagents}; and incomplete pathology correlation, as demonstrated by Wang et al. \cite{wang2024beyond}, where agent systems showed weaker alignment with histopathological grades than radiologists.

The embodied agent framework proposed by Qiu et al. \cite{qiu2025embodied} suggests a promising direction. This framework combines three key elements: Perceptual Specialization, with dedicated convolutional agents for nodule margin analysis; Knowledge Anchoring, with on-demand retrieval from NCCN guidelines; and Uncertainty-Aware Collaboration, with confidence-weighted voting among subspecialty agents.

\begin{figure}[t]
\centering
\includegraphics[width=1\columnwidth]{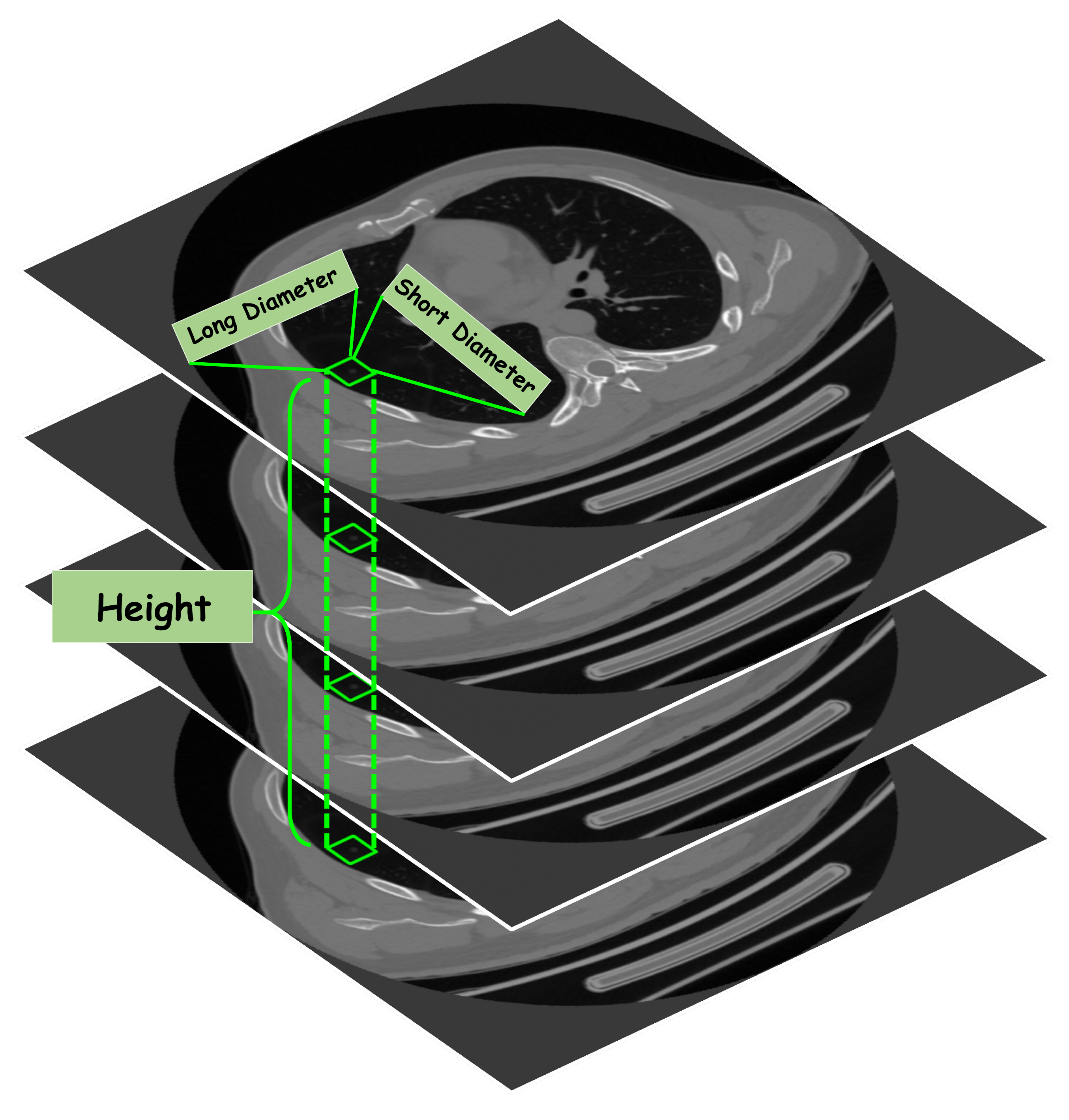}
\caption{Diagram of Measuring Lung Nodule Size.}
\label{fig:3}
\end{figure}

\begin{figure*}[t]
\centering
\includegraphics[width=1.8\columnwidth]{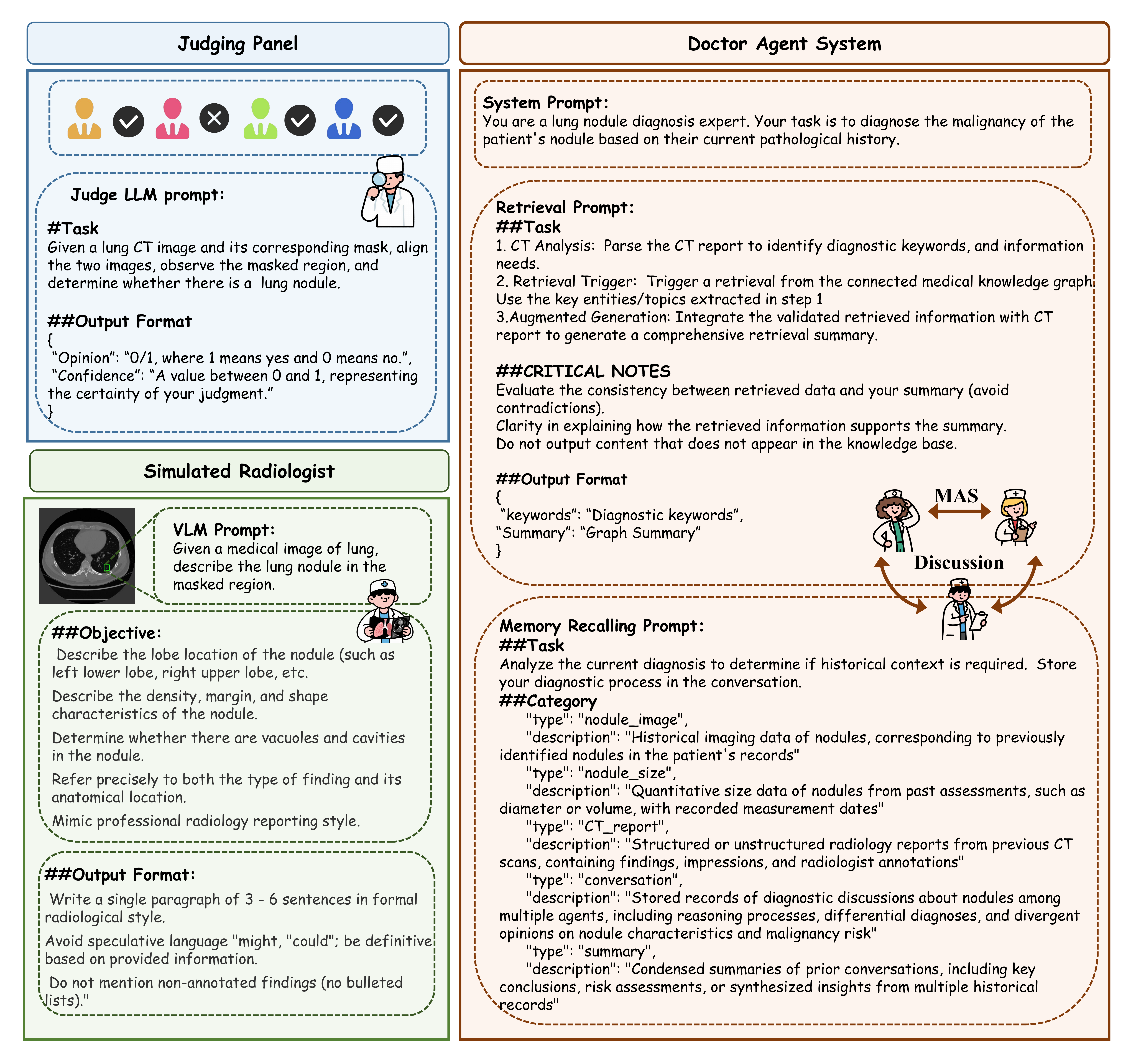}
\caption{Prompts for three main modules.}
\label{fig:1}
\end{figure*}

\begin{figure*}[t]
\centering
\includegraphics[width=2\columnwidth]{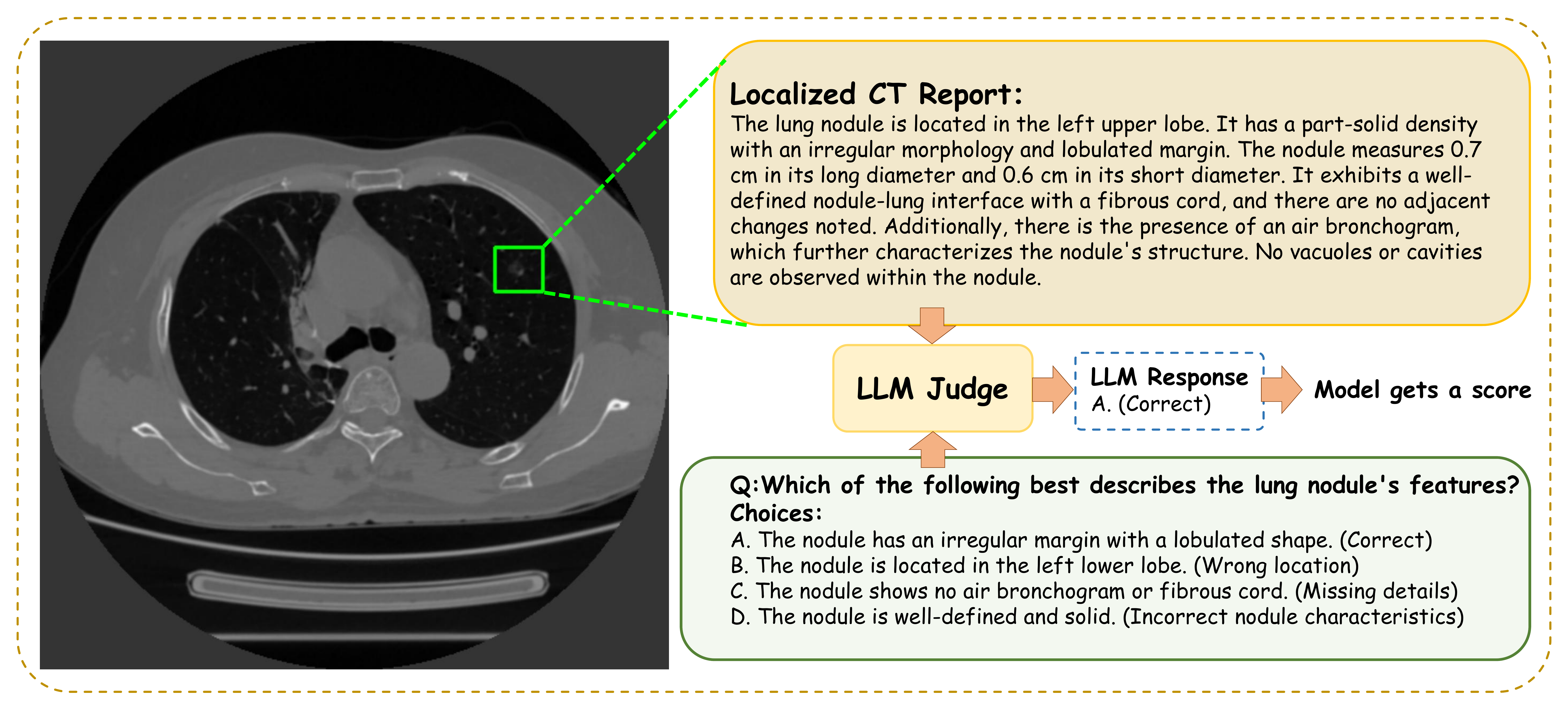}
\caption{\textbf{LungDLC evaluation process}: a question-answering task is constructed based on domain-specific attributes; an evaluator driven by a large language model (LLM) assigns correctness labels to the question-answering results, and the model receives a corresponding score when the generated description matches the ground-truth semantic attributes. This framework provides a feasible path for reference-free benchmarking of fine-grained regional description performance in the field of medical image understanding. It is worth noting that although the widely used LLM score and LungDLC-score both rely on the LLM evaluation mechanism, the former is more effective in natural vision-language scenarios, while the latter is a tailored evaluation metric for specific tasks in the medical domain.}
\label{fig:2}
\end{figure*}

\section{Technical Detail}

\subsection{Nodule Size Details}

To accurately measure the size of a lung nodule in a 3D CT scan using a lung nodule mask, begin by identifying the axial slice that shows the nodule's largest cross-sectional area. On this slice, measure the long diameter, which is the widest part of the nodule, and the short diameter, perpendicular to the long diameter. Next, monitor the nodule across consecutive axial slices to observe any changes in its appearance. The height is determined by the extent of the nodule along the z-axis. These three dimensions—long diameter, short diameter, and height—together define the size of the lung nodule. The nodule's volume can then be estimated using the formula:
\begin{equation}
V = \frac{1}{6} \cdot \text{long diameter} \cdot \text{short diameter} \cdot \text{height}.
\end{equation}

\subsection{Prompt Details}
As illustrated in \textbf{Figure \ref{fig:1}}, the prompt templates collectively outline the workflow of a multi-agent system developed for diagnosing lung nodules, from initial screening to final decision-making. Initially, the Judge LLM conducts an image-level assessment, examining a lung CT image and its mask to ascertain the presence of a nodule in the masked area. It provides its evaluation as an Opinion'' along with a Confidence'' value, offering a calibrated estimate of how reliable this binary decision is.

Following this, the Simulated Radiologist Prompt is utilized to offer a detailed description of the nodule in a professional radiological style. This prompt instructs the agent to detail the nodule's lobe location, density, margin, and shape, as well as identify any vacuoles or cavities and relate them to standard radiological reporting practice. The response should be a concise paragraph of 3--6 sentences, written formally and without speculative language, so that it can be directly read or post-processed as a structured radiology report.

Finally, the Doctor Agent System employs multiple prompts to deliver a comprehensive diagnosis based on both the current case and accumulated prior knowledge. The Retrieval Prompt examines the CT report to pinpoint diagnostic keywords and integrates relevant data from a medical knowledge graph into a coherent summary that explains how external evidence supports the case. Simultaneously, the Memory Recalling Prompt evaluates whether historical context is necessary for the current diagnosis and records the diagnostic process for future reuse. This historical context encompasses categories such as ``nodule image'', ``nodule size'', and ``CT report'', as well as prior diagnostic discussions.

\subsection{Doctor Agent System Details}

\begin{algorithm}[H]
\caption{DAS Workflow}
\label{alg:das}
\begin{algorithmic}[1]  
\REQUIRE CT volume $V$
\ENSURE Final diagnosis $\mathcal{FD}$

\STATE Construct medical knowledge graph $\mathcal{G}$ from documents $\mathcal{D}$
\STATE Generate summaries $\mathcal{S}$ from $\mathcal{G}$
\FOR{$i = 1$ to $K$}
    \STATE Initialize $O_i^{(1)} \gets \text{Agent}_i(I, \text{Report})$
\ENDFOR
\WHILE{consensus not reached}
    \FOR{each agent $i$}
        \STATE $O_i^{(t)} \gets \text{Revise}\big(O_i^{(t-1)}, \{O_j^{(t-1)}\}_{j \ne i}\big)$
    \ENDFOR
    \STATE Summarizer combines $\{O_i^{(t)}\}$ into intermediate result
\ENDWHILE
\STATE $\mathcal{FD} \gets \text{Consensus}\big(\text{Summarizer}(\{O_i^{(t^*)}\})\big)$

\STATE \textbf{Memory module:}
\STATE Store nodule images, masks, size metrics
\STATE Store CT reports $\mathcal{O}_{\text{vlm}}$
\STATE Store multi-agent conversations and summaries
\end{algorithmic}
\end{algorithm}

\section{Experiment Details}

\subsection{LungDLC-score}
As depicted in \textbf{Figure \ref{fig:1}}, we designed a set of clinically relevant yes/no questions for each nodule, using dataset annotations to evaluate the descriptions generated. The average correctness rate from these questions produces the LungDLC-score, a dependable measure for assessing how accurately reports capture key clinical features and morphology.

\subsection{Result Visualization}

As illustrated in \textbf{Table \ref{tab:1}}, while other approaches sometimes correctly identify certain features, they often fail to describe all the characteristics of a nodule accurately, occasionally producing incorrect details. In contrast, LungNoduleAgent excels in accurately describing the local features of pulmonary nodules, offering precise supporting evidence. This highlights its superior diagnostic accuracy in interpreting pulmonary nodules.

\begin{table*}[h!]
\centering
\begin{tabular}{p{0.15\linewidth} p{0.2\linewidth} p{0.6\linewidth}}
\toprule
\textbf{Nodule Image} & \multicolumn{2}{c}{%
  \begin{minipage}[c]{0.2\linewidth}
    \centering
    \includegraphics[width=\linewidth]{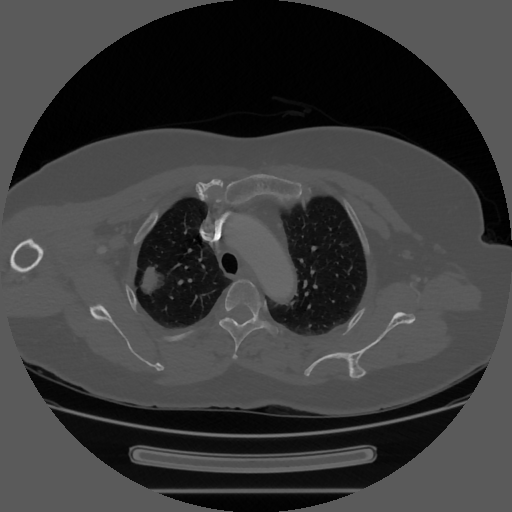}\\
    Image
  \end{minipage}%
  \qquad
  \begin{minipage}[c]{0.2\linewidth}
    \centering
    \includegraphics[width=\linewidth]{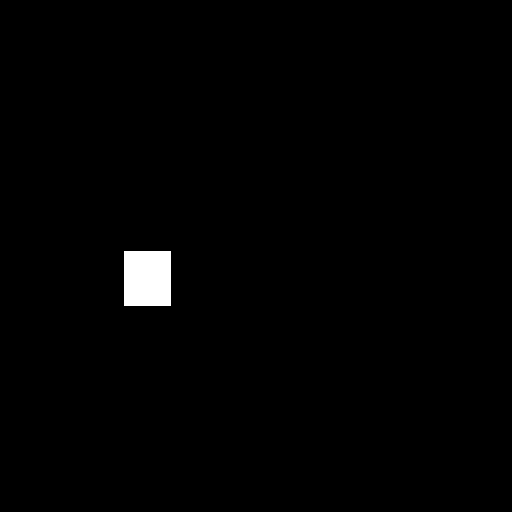}\\
    Mask
  \end{minipage}%
} \\
\midrule

\textbf{GPT-4o} & \multicolumn{2}{p{0.8\linewidth}}{The masked CT axial image reveals a solitary pulmonary nodule located in the\textbf{ right upper lobe} of the lung. The nodule is \underline{well-circumscribed with smooth margins} and demonstrates soft-tissue attenuation, consistent with a solid density. \textbf{The shape is round to oval}, \underline{with no evidence of spiculation or lobulation.} There is \textbf{no identifiable cavitation or internal vacuolization within the lesion.} The surrounding lung parenchyma appears unremarkable, with no adjacent consolidation or pleural reaction.} \\
\midrule

\textbf{Claude 3.7 Sonnet} & \multicolumn{2}{p{0.8\linewidth}}{Based on the axial CT image, there is a well-defined pulmonary nodule located in the \underline{left lower} lobe. The nodule \underline{demonstrates solid density with smooth, well-circumscribed margins} and maintains \textbf{a round to oval morphology.} \textbf{No internal cavitation or vacuolation is evident within the lesion.} The nodule appears \underline{homogeneous in attenuation without apparent calcification or fat density components.} The surrounding lung parenchyma shows no immediate inflammatory changes or architectural distortion adjacent to the nodule.} \\
\midrule

\textbf{InternVL} & \multicolumn{2}{p{0.8\linewidth}}{The image reveals a solitary pulmonary nodule located in the left upper lobe. The nodule exhibits \underline{a well-defined margin} and a \textbf{predominantly rounded shape}. \underline{Its density appears homogeneous}, contrasting with surrounding lung tissue. No cavities or vacuoles are observed within the nodule. The anatomical positioning suggests a discrete lesion, warranting further evaluation.} \\
\midrule

\textbf{LLaVA} & \multicolumn{2}{p{0.8\linewidth}}{The lung nodule is located in the \underline{upper lobe}. The nodule exhibits a \underline{solid}, \textbf{round shape} with a thin, \underline{poorly defined margin}. Its density is slightly increased compared to surrounding lung parenchyma. \textbf{No visible vacuoles or cavities are present within the nodule}. The red mask suggests this may be a preliminary diagnosis or visual aid for identification. Overall, it appears as a well-defined, solid mass in the upper lung lobe.} \\
\midrule

\textbf{Medgemma} & \multicolumn{2}{p{0.8\linewidth}}{The masked region shows \underline{a solitary pulmonary nodule in the left upper lobe.} The nodule is \underline{approximately 1.5 cm in diameter with solid density and smooth, well-defined margins.} \textbf{The shape is roughly round without evidence of internal vacuoles or cavitation.}} \\
\midrule

\textbf{Qwen2.5-VL} & \multicolumn{2}{p{0.8\linewidth}}{The lung nodule is located in the \underline{right lower lobe}, appearing as a well-defined, homogenous, slightly hypodense mass with smooth margins. \textbf{It has a round to oval shape without visible vacuoles or cavities.} \underline{The findings are consistent with a benign lesion, though further evaluation is recommended.}} \\
\midrule

\textbf{MedR1} & \multicolumn{2}{p{0.8\linewidth}}{The axial CT scan shows a well-defined, oval-shaped nodule in the right upper lobe. The nodule has \underline{homogenous density without cavitation or air bronchograms.} \underline{Margins are smooth and well-defined} \underline{without irregularities.} \textbf{No signs of vacuoles or cystic spaces} are present. The location is typical for lung nodules. No associated abnormalities like pleural effusion are seen. Findings suggest a benign nodule, \underline{though further imaging/biopsy is recommended.}} \\
\midrule

\textbf{LungNoduleAgent} & \multicolumn{2}{p{0.8\linewidth}}{The lung nodule is located in the \textbf{right upper lobe}. It appears to have a \textbf{part-solid density} with a\textbf{round or oval shape and lobulated margins}. The nodule demonstrates \textbf{well-defined spiculation at the lung nodule interface}. There are \textbf{no air bronchograms or cavities} present within the nodule. Additionally, there is an \textbf{adjacent vascular convergence sign and pleural indentation} noted.} \\

\bottomrule
\end{tabular}
\caption{Visual comparison on CT Report Generation task. \textbf{bolded} for correct and \underline{underline} for wrong.}
\label{tab:1}
\end{table*}

\end{document}